\documentclass[runningheads]{llncs}

\usepackage[mobile]{eccv}

\usepackage{eccvabbrv}

\usepackage{graphicx}
\usepackage{booktabs}
\usepackage{amsmath}
\usepackage{caption}
\usepackage{float}
\usepackage[linesnumbered,ruled,vlined]{algorithm2e}
\usepackage{enumitem}
\usepackage{multirow}

\usepackage[accsupp]{axessibility}  

\usepackage{hyperref}

\usepackage{orcidlink}

\begin{document}

\title{Stokes-Informed Diffusion for Robust Linear Polarization Estimation} 

\titlerunning{GenPolar}

\author{Yidong Luo\inst{1,2}\thanks{Equal contribution. \quad $^\dagger$ Corresponding authors.}\orcidlink{0000-0002-9665-6471} \and
Chenggong Li\inst{3}\protect\footnotemark[1]\orcidlink{0009-0003-9955-642X} \and Yuchao Feng\inst{4}\orcidlink{0000-0001-6097-3806} \and Boxin Shi\inst{5}\orcidlink{0000-0001-6749-0364} \and \\ Junchao Zhang\inst{3}$^\dagger$\orcidlink{0000-0003-2243-0012} \and Xin Yuan\inst{2}$^\dagger$\orcidlink{0000-0002-8311-7524}}

\authorrunning{Y. Luo, C. Li et al.}

\institute{Zhejiang University, Hangzhou, China \and
School of Engineering, Westlake University, Hangzhou, China \\
\email{\{luoyidong, xyuan\}@westlake.edu.cn}
\and School of Automation, Central South University, Changsha, China \\
\email{\{244603040, junchaozhang\}@csu.edu.cn}
\and College of Computer Science and Technology, Zhejiang University of Technology, Hangzhou, China
\and School of Computer Science, Peking University, Beijing, China}

\maketitle

\begin{abstract}
Polarization cues benefit applications such as material detection and de-reflection, yet acquiring them typically requires dedicated hardware. This motivates us to estimate the linear polarization from a single RGB image. However, the task is inherently ill-posed, with the Angle of Polarization (AoP) becoming particularly unstable in weak-polarization regions, where the polarimetric signal is overwhelmed by noise, leading to erratic angle estimates. To address these limitations, we propose GenPolar, a Stokes-informed diffusion framework grounded in the Mueller formalism from an intensity observation. Specifically, GenPolar predicts channel-wise linear Stokes components $(S_{1},S_{2})$ from intensity $S_0$, from which degree of linear polarization (DoLP) and AoP are analytically derived; AoP is further supervised with an observability-aware loss. In addition, to enable efficient and high-fidelity inference, we adopt a two-stage training strategy. Firstly, a multi-step conditional diffusion model is trained with a physics-based loss. Subsequently, we distill it into a one-step generator, which further supports stable Low-Rank Adaptation (LoRA) of the VAE encoder to mitigate domain-specific autoencoding bias. Extensive experiments across rotating-polarizer, division-of-focal-plane, and hybrid datasets demonstrate that GenPolar achieves state-of-the-art performance in both DoLP fidelity and AoP stability. Crucially, these improvements translate to significant and consistent gains in downstream applications, including material detection and de-reflection.

\end{abstract}

\section{Introduction}
\label{sec:intro}

\begin{figure}[tb]
  \centering
  \includegraphics[width=\textwidth]{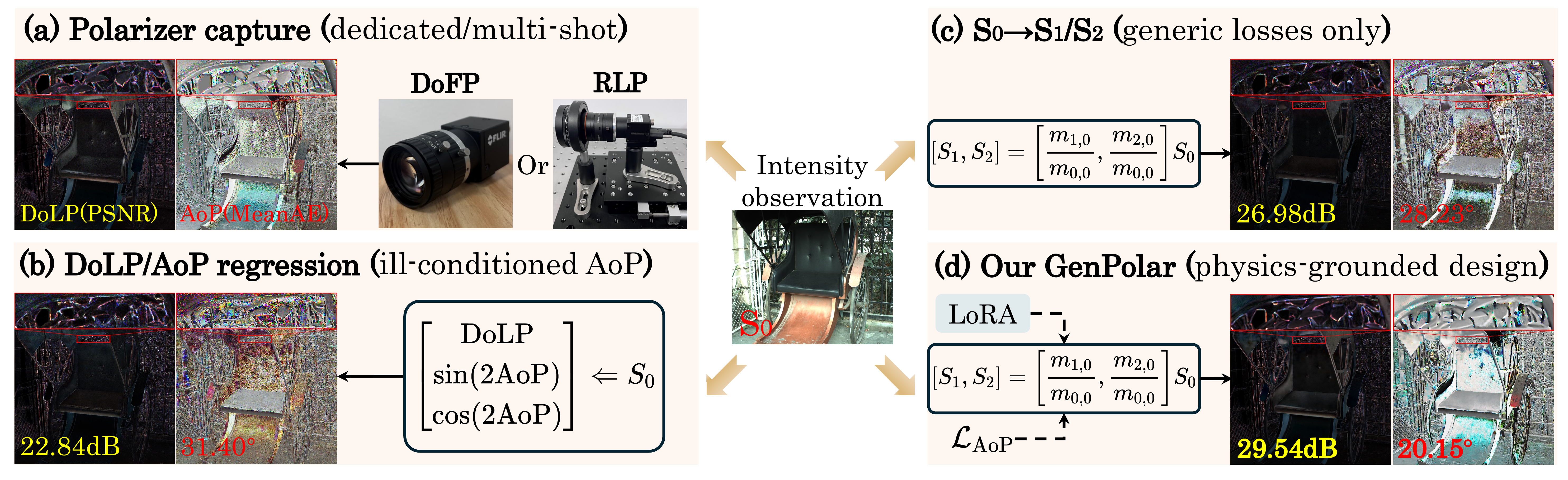}
  \caption{(a) DoFP/RLP capture requires specialized hardware. (b) Direct DoLP/AoP regression is unstable in weakly polarized regions. (c) Predicting $(S_1,S_2)$ from $S_0$ offers a Mueller-grounded alternative. (d) GenPolar further uses AoP observability-aware supervision and LoRA adaptation for more coherent cues.}
  \label{fig:1}
\end{figure}

Polarization reveals geometric and material cues that are largely imperceptible to standard RGB imaging, supporting applications such as material detection and reflection separation~\cite{collett1992polarized}. However, capturing polarization typically depends on dedicated hardware, i.e., division-of-focal-plane polarimeters DoFP~\cite{zhang2016image}, or multi-shot capture with a rotating linear polarizer (RLP; Fig.~\ref{fig:1}(a)), incurring additional cost and constraining deployment. These limitations motivate single-image polarization estimation from an intensity observation, with the goal of recovering cues that are not only visually consistent but also physically grounded and reliable for downstream reasoning.

Notably, when the observation is an RGB intensity image, polarization is not a single scalar property but exhibits wavelength-dependent behavior~\cite{azzam1978ellipsometry}. In many common formation processes, the degree of linear polarization (DoLP) varies with wavelength due to wavelength-dependent diattenuation and dispersion, whereas angle of polarization (AoP) is often less sensitive to wavelength under typical surface-reflection settings~\cite{chenault1993measurements}. Consequently, for an RGB observation, the polarization parameters should be channel-wise estimated to preserve physically meaningful spectral differences in polarization strength.

However, an RGB intensity observation collapses the underlying polarization formation into a severely incomplete measurement, making single-image polarization estimation fundamentally ill-posed: the scene-dependent polarization transfer is unknown and varies spatially with material, geometry, and illumination. In addition, AoP is \emph{not uniformly observable}~\cite{wolff2002constraining,chipman2018polarized}: when DoLP is small, AoP becomes ill-conditioned and even minor perturbations can induce large angular fluctuations, which explains the speckled AoP maps produced by direct DoLP/AoP regression in Fig.~\ref{fig:1}(b). A seemingly safer alternative is to predict Stokes components and derive DoLP/AoP analytically, yet optimizing $(S_1,S_2)$ in latent diffusion model~\cite{rombach2022high} with generic reconstruction losses can still yield unstable cues (Fig.~\ref{fig:1}(c)), partly because Stokes fields follow different statistics from natural RGB images and off-the-shelf latent VAEs introduce structured autoencoding bias. These observations motivate us to seek a physically grounded and spectrally faithful $S_0\!\rightarrow\!(S_1,S_2)$ formulation, together with observability-aware AoP supervision and stable latent-domain adaptation.

In this paper, we propose \textbf{GenPolar}, a Stokes-informed diffusion framework grounded in the Mueller formalism~\cite{chenault1993measurements} for single-image linear polarization estimation. We cast the task as channel-wise $S_{0,\lambda}\!\rightarrow\!(S_{1,\lambda},S_{2,\lambda})$, preserving the spectral dependence of polarization strength, and derive DoLP/AoP as standard cues. To avoid fitting ill-conditioned angles, we supervise AoP only on a DoLP-thresholded observability mask. Finally, we adopt a two-stage training strategy: a physics-supervised multi-step diffusion base model, followed by one-step distillation~\cite{yin2024one} that enables stable LoRA adaptation~\cite{hu2021lora} of the VAE encoder to reduce polarization-domain autoencoding bias. Experiments on RLP, DoFP, and hybrid datasets show consistent gains in DoLP fidelity and AoP stability, and we further verify their utility on material detection and polarization de-reflection. Our contributions are summarized as:
\begin{itemize} 
    \item A Mueller-grounded, RGB channel-wise $S_0\!\rightarrow\!(S_1,S_2)$ formulation with ob-servability-aware AoP supervision.
    \item A two-stage diffusion training with one-step distillation enabling stable encoder LoRA adaptation.
    \item Cross-modality evaluation, with downstream validation on material detection and polarization de-reflection.
\end{itemize}

\section{Related Work}

\textbf{Hardware for polarization acquisition.}
Polarization is commonly represented by Stokes parameters, from which DoLP and AoP are computed for geometry and material reasoning, often focusing on linear polarization $(S_1,S_2)$. Measuring polarization requires optical modulation that maps polarization states to intensity observations. Typical systems include DoFP polarimeters with micro-polarizer mosaics (single-shot but spatially multiplexed, requiring polarimetric demosaicking) and rotating-analyzer capture with a rotating linear polarizer (high fidelity but multi-shot and motion-sensitive)~\cite{chipman2018polarized}. Other single-shot designs include division-of-amplitude~\cite{azzam1982division} and division-of-aperture polarimeters~\cite{pezzaniti2005division} that split the beam/sub-aperture into multiple analyzer channels, as well as liquid-crystal modulation~\cite{coleman2003polarization} for programmable polarization analysis. These designs trade off optical complexity, calibration burden and reconstruction difficulty, motivating polarization estimation from standard RGB observations.

\noindent\textbf{Polarization restoration and downstream tasks.}
Since many acquisition pipelines multiplex polarization in space or time, a substantial body of work studies polarization restoration and reconstruction, including DoFP demosaicking~\cite{luo2024learning,zhou2025pidsr,li2025demosaicking,li2026pugdiff}, denoising~\cite{luo2024learning,li2023polarized,li2023joint} and so on, often exploiting correlations across analyzer angles and Stokes-domain constraints. Polarization cues are then used in downstream pipelines such as material detection~\cite{lyu2024sfpuel,xiang2021polarization,kalra2020deep,terrier2008segmentation,mei2022glass}, de-reflection~\cite{yao2025polarfree,lei2020polarized,lyu2019reflection}, normal estimation~\cite{li2024neisf,lei2022shape}, low-light enhancement~\cite{hu2020iplnet,zhou2023polarization,lu2024polarization} and dehazing~\cite{yang2025novel,sun2025self}. In such applications, AoP can be highly informative yet unstable under weak polarization, so practical systems benefit from explicitly handling angle uncertainty rather than assuming uniformly reliable AoP.

\noindent\textbf{Polarization estimation from RGB.}
Recent work formulates polarization estimation as predicting polarization information directly from an RGB input. The benchmark study RGB-to-Polarization Estimation~\cite{lin2025rgb2pol} establishes RGB-to-polarization as a task and evaluates diverse model families, treating the RGB image as the total intensity ($S_0$) and estimating polarization components in a Stokes parameterization. In parallel, PolarAnything~\cite{zhang2025polaranything} explores diffusion-based polarimetric synthesis from a single RGB image, and shows that representing polarization with compact cues (e.g., DoLP/AoP) can facilitate generative modeling. Overall, these studies highlight the promise of strong priors for plausible polarization synthesis, while leaving open how to obtain AoP/DoLP that are consistently usable for downstream inference, especially under weak polarization. In contrast, we adopt an observability-aware, physics-grounded Stokes estimation paradigm that targets more stable AoP/DoLP, and assess its practical utility through downstream stress tests.

\section{Methodology}

\subsection{Physical plausibility of $S_{0,\lambda}$$\rightarrow$$(S_{1,\lambda},S_{2,\lambda})$}

We represent polarization by the Stokes vector $\mathbf{s}=[S_{0},S_{1},S_{2},S_3]^\top$,
where $S_{0}$ is the total intensity and $(S_{1},S_{2})$ encode linear polarization; we assume negligible circular polarization ($S_3\approx0$).
Given four analyzer-angle measurements $I_{0^\circ},I_{45^\circ},I_{90^\circ},I_{135^\circ}$, the Stokes components are computed as
\begin{equation}
\textstyle
 S_{0} = \frac{I_{0^\circ}+I_{45^\circ}+I_{90^\circ}+I_{135^\circ}}{2}, \quad
 S_{1} = I_{0^\circ}-I_{90^\circ},\quad
 S_{2} = I_{45^\circ}-I_{135^\circ}.
\label{eq:1}
\end{equation}
DoLP and AoP are then computed as standard cues:
\begin{equation}
\textstyle
\mathrm{DoLP}=\frac{\sqrt{S_{1}^2+S_{2}^2}}{S_{0}},\qquad
\mathrm{AoP}=\frac{1}{2}\operatorname{atan2}({S_2},{S_1}).
\label{eq:2}
\end{equation}

\begin{figure}[tb]
  \centering
  \includegraphics[width=\textwidth]{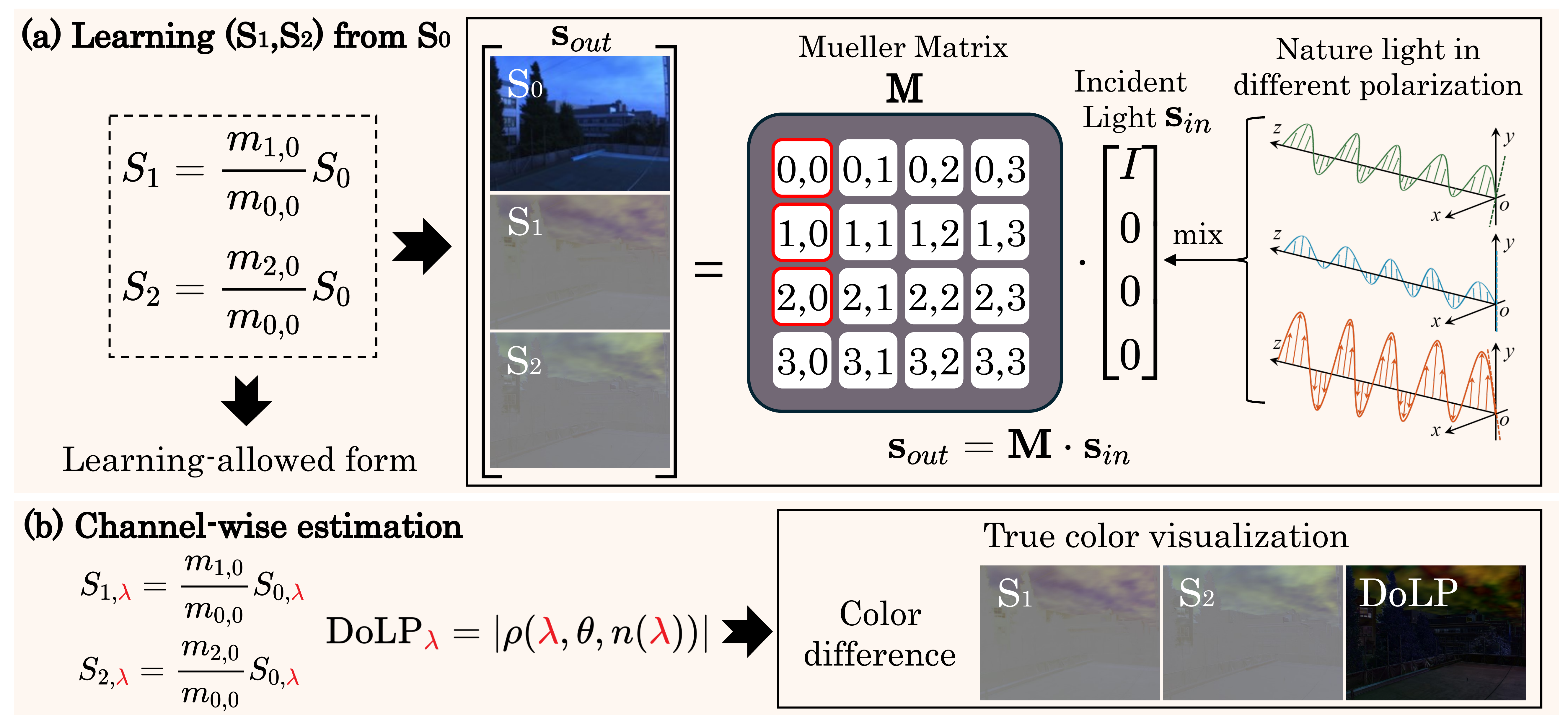}
  \caption{Physical motivation for the Stokes-estimation formulation. (a) Under an unpolarized illumination approximation, Mueller transport implies an effective $S_0$$\rightarrow$$(S_1,S_2)$ relation, making intensity-to-linear-polarization estimation physically plausible. (b) Polarization strength is spectral, motivating channel-wise estimation.}
  \label{fig:2}
\end{figure}

The task definition is summarized in Fig.~\ref{fig:2}. The physical plausibility of inferring nonzero $(S_1,S_2)$ from intensity-only input can be viewed through Mueller transport~\cite{chenault1993measurements}. In natural scenes, the incident illumination can be approximated as unpolarized, since it is a mixture of many polarization states with random orientations. For incident light $\textbf{s}_{in}=[I,0,0,0]^\top$, the outgoing Stokes $\textbf{s}_{out}$ satisfies
\begin{equation}
\small
    \textbf{s}_{out}= 
    \begin{bmatrix}
S_{0} \\ S_{1} \\ S_{2} \\ S_{3}
\end{bmatrix}=
    \textbf{M}\cdot\textbf{s}_{in}=
\begin{bmatrix}
 m_{0,0} & m_{0,1} & m_{0,2} & m_{0,3}\\
 m_{1,0} & m_{1,1} & m_{1,2} & m_{1,3}\\
 m_{2,0} & m_{2,1} & m_{2,2} & m_{2,3}\\
 m_{3,0} & m_{3,1} & m_{3,2} & m_{3,3}
\end{bmatrix}
\begin{bmatrix}
 I\\
 0\\
 0\\
 0
\end{bmatrix}
=
I
\begin{bmatrix}
m_{0,0} \\ m_{1,0} \\ m_{2,0} \\ m_{3,0}
\end{bmatrix}.
\label{eq:mueller}
\end{equation}
Here $\mathbf{M}$ is the scene Mueller matrix, and $S_3\approx0$ in linear polarization. $m_{0,0}$ measures the intensity throughput, while $m_{1,0}$ and $m_{2,0}$ quantify how unpolarized intensity is converted into linear polarization along the $0^\circ/90^\circ$ and $45^\circ/135^\circ$ bases, respectively, indicating that nonzero $m_{1,0}$ or $m_{2,0}$ can produce linear polarization. Since $S_{0}=m_{0,0}I$, $S_{1}=m_{1,0}I$, and $S_{2}=m_{2,0}I$ in Eq.~\eqref{eq:mueller}, the linear polarization components are coupled to the intensity via the Mueller parameters. We therefore define the task as estimating $(S_{1},S_{2})$ from $S_{0}$. Under the standard assumption of unpolarized incident illumination, Mueller transport implies
\begin{equation}
\textstyle
S_{1}=\frac{m_{1,0}}{m_{0,0}}S_{0}, \qquad
S_{2}=\frac{m_{2,0}}{m_{0,0}}S_{0},
\label{eq:s0_s1s2}
\end{equation}
\noindent where the normalized ratios $\frac{m_{1,0}}{m_{0,0}}$ and $\frac{m_{2,0}}{m_{0,0}}$ can be viewed as components of an effective diattenuation, which varies with local material, geometry, and illumination. Importantly, Eq.~\eqref{eq:s0_s1s2} is a physics-derived structural constraint from Mueller transport under unpolarized illumination: it shows that $(S_{1},S_{2})$ can be written as a spatially varying linear transfer of $S_{0}$, with coefficients determined by scene polarimetric properties. It provides a physically plausible parameterization that our model learns under physics supervision. Together with $S_{0}$, the predicted $(S_{1},S_{2})$ allows us to compute $\mathrm{DoLP}$ and $\mathrm{AoP}$ via Eq.~\eqref{eq:2}. In scenarios with partially polarized illumination or complex propagation (e.g., skylight), Eq.~\eqref{eq:s0_s1s2} should be understood in an effective sense.

A key formation property motivating our RGB channel-wise prediction is that polarization strength depends on wavelength and incidence geometry.
For a broad class of diattenuation/Fresnel effects~\cite{wolff2002constraining}, linear Stokes components admit the amplitude-orientation form
\begin{equation}
\textstyle
\begin{bmatrix} S_{1,\lambda}, S_{2,\lambda}\end{bmatrix}
=\rho(\lambda,\theta,n(\lambda))\,S_{0,\lambda}
\begin{bmatrix}\cos 2\psi, \sin 2\psi\end{bmatrix},
\label{eq:fresnel}
\end{equation}
where \( \rho(\lambda, \theta, n(\lambda)) \) captures the polarization strength and depends on wavelength \( \lambda \), the local incidence angle \( \theta \), and the refractive index \( n(\lambda) \) of the material, which accounts for wavelength-dependent diattenuation and dispersion. The angle of polarization \( \psi \) reflects the orientation of the plane of incidence, primarily determined by macroscopic geometry, such as surface normals and viewing directions. This decomposition yields $\mathrm{DoLP}_\lambda=|\rho(\lambda,\theta,n(\lambda))|$ and $\mathrm{AoP}_\lambda=\psi$.
Thus DoLP is inherently wavelength-dependent, motivating estimation of $(S_{1,\lambda},S_{2,\lambda})$ for all RGB channels rather than collapsing polarization to grayscale. 
\begin{figure}[tb]
  \centering
  \includegraphics[width=0.99\textwidth]{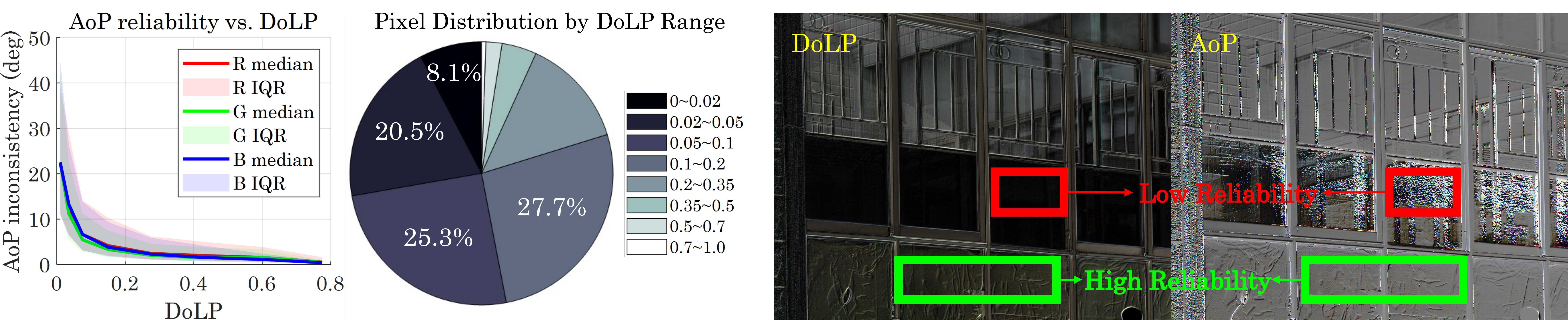}
  \caption{AoP reliability analysis across public datasets. The line plot shows the decreasing trend of cross-channel error with increasing DoLP. The pie chart illustrates the pixel-wise DoLP distribution across all images in public datasets. True-color visualizations of DoLP and AoP clearly reveal pronounced noise in weakly polarized regions.}
  \label{fig:3}
\end{figure}

A seemingly simpler alternative is to regress ($\mathrm{DoLP}_\lambda$, $\cos(2\mathrm{AoP}_\lambda)$, $\sin(2\mathrm{AoP}_\lambda)$) directly from $S_{0,\lambda}$. Yet this discards the explicit Stokes-level structure in Eq.~\eqref{eq:s0_s1s2} and does not resolve the instability of AoP under weak polarization. We quantify this instability on public datasets by cross-channel AoP coherence, defined as the mean angular deviation of one channel relative to the other two, which serves as a practical reliability proxy. As shown in Fig.~\ref{fig:3}, this deviation is much larger at low DoLP and decreases rapidly as DoLP increases. These observations motivate an observability-aware objective and evaluation that explicitly account for where polarization cues are reliable.

Overall, we implement $S_{0,\lambda}\rightarrow(S_{1,\lambda},S_{2,\lambda})$ using a pretrained stable diffusion model and fine-tune it on polarization data, letting the generative prior regularize this ill-posed inference while learning the spatially varying polarization transfer implied by Eq.~\eqref{eq:s0_s1s2}. Moreover, the pretrained model provides a rich representation: its features encode appearance cues correlated with geometry, material, and illumination factors that drive the transfer in Eq.~\eqref{eq:s0_s1s2}.

\subsection{Stokes-Informed Stable Diffusion}
\begin{figure}[t]
  \centering
  \includegraphics[width=0.99\textwidth]{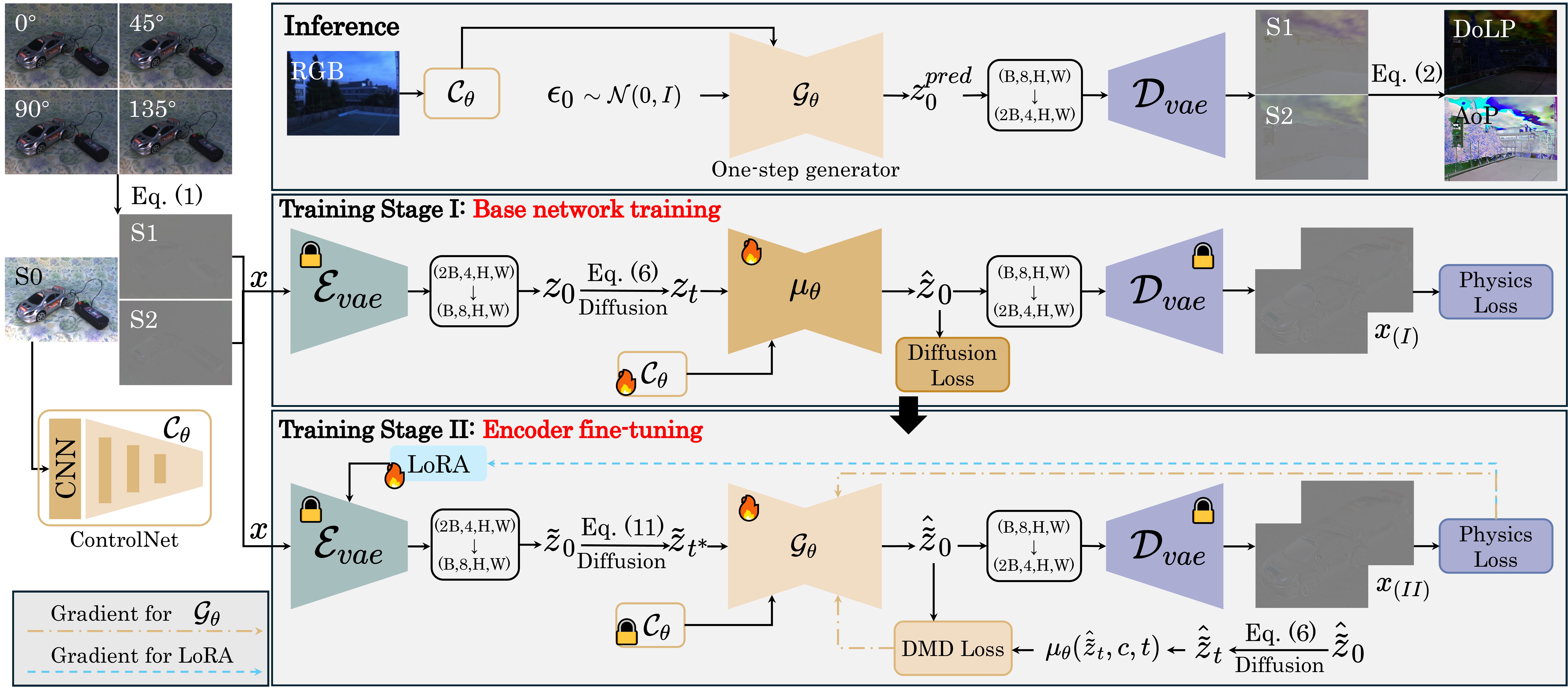}
  \caption{Overview of GenPolar. Given RGB intensity proxy $S_{0}$, GenPolar predicts $(S_{1},S_{2})$ and derives DoLP/AoP. Stage I trains a multi-step diffusion model with physics supervision, and Stage II distills it into a one-step generator with VAE-encoder LoRA.}
  \label{fig:4}
\end{figure}

We propose a Stokes-Informed Stable Diffusion framework for generating polarization components $(S_{1,\lambda},S_{2,\lambda})$ from RGB intensity proxy proportional to $S_{0,\lambda}$. The overall pipeline is shown in Fig.~\ref{fig:4}. Our method operates in a clear inference process, followed by two training stages. At inference time, the model receives an RGB intensity proxy $c$ proportional to $S_{0,\lambda}$ and Gaussian noise $\epsilon_0$. The one-step generator $\mathcal{G}_\theta$ predicts the clean latent $z_0^{pred}=\mathcal{G}_\theta(\epsilon_0, c, t^*)$, where $t^*$ is the fixed timestep corresponding to the maximum noise level. This latent $z_0^{pred}$ is then decoded by $\mathcal{D}_{vae}$ to obtain $(\hat S_{1,\lambda}, \hat S_{2,\lambda})$.

\begin{figure}[tb]
  \centering
  \includegraphics[width=\textwidth]{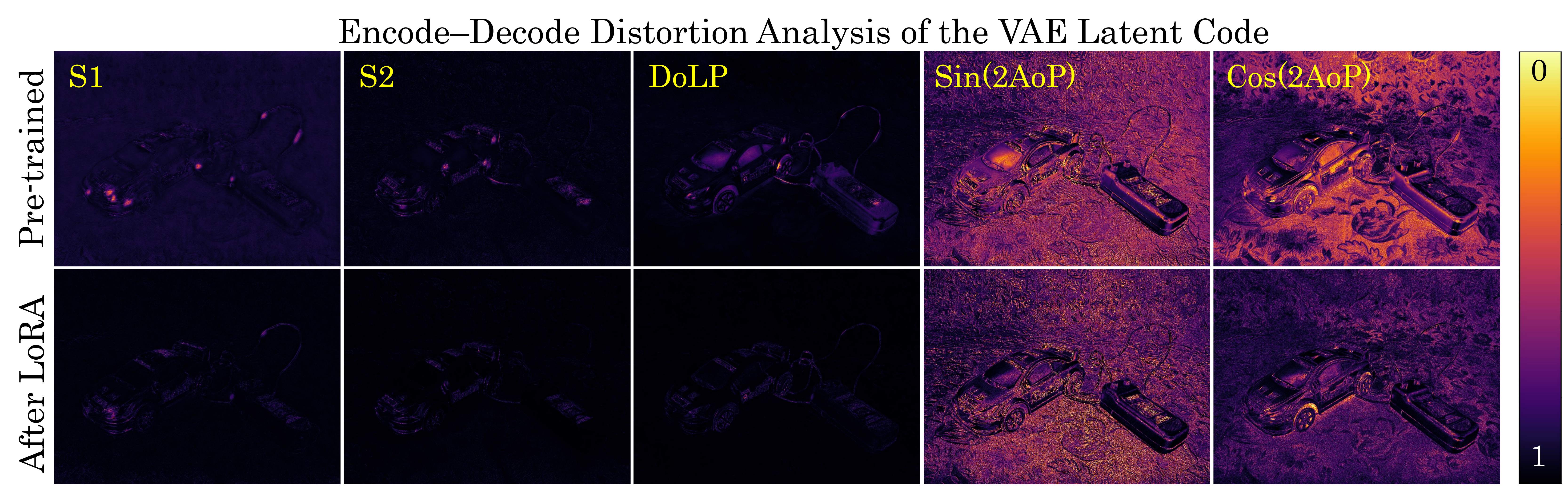}
  \caption{The error map is significantly reduced after LoRA adaptation, demonstrating its effectiveness in improving the reconstruction of polarization targets.}
  \label{fig:5}
\end{figure}

Directly encoding and decoding polarization targets with the pretrained VAE yields noticeable errors (Fig.~\ref{fig:5}), indicating a mismatch between polarization maps and the pre-trained latent representation. We therefore insert LoRA adapters~\cite{hu2021lora} into the VAE encoder $\mathcal{E}_{vae}$ to mitigate this autoencoding bias, while keeping the decoder fixed to maintain a consistent output parameterization. However, stably adapting $\mathcal{E}_{vae}$ requires end-to-end optimization, which is impractical to backpropagate through long multi-step diffusion trajectories. This motivates our two-stage training: ($i$) Stage I trains the multi-step diffusion backbone $\mu_\theta$ and ControlNet~\cite{zhang2023adding} $\mathcal{C}_{\theta}$ while keeping the VAE frozen. ($ii$) Stage II distills the multi-step diffusion process into a one-step generator $\mathcal{G}_\theta$ and adapts the $\mathcal{E}_{vae}$ via LoRA. As evidenced in Fig.~\ref{fig:5}, the LoRA-adapted encoder substantially reduces the reconstruction error, validating our two-stage training scheme.

\noindent\textbf{Training stage I}: In Stage I, we start with a pretrained VAE encoder that produces a 4-channel latent representation. Since $(S_{1,\lambda},S_{2,\lambda})$ form a 6-channel target (denoting as $x=[S_{1,\lambda},S_{2,\lambda}] \in \mathbb{R}^{6\times H\times W}$), we encode the 3-channel $S_{1,\lambda}$ and $S_{2,\lambda}$ separately using the same encoder, then concatenate the resulting latents into an 8-channel tensor $z_0 \in \mathbb{R}^{8 \times \frac{H}{8} \times \frac{W}{8}}$. The noisy latent $z_t$ is given by 
\begin{equation}
    z_t=\alpha_t z_0 + \sigma_t \epsilon,
\label{eq:diffusion}
\end{equation}
where $\alpha_t$ and $\sigma_t$ are determined by the noise schedule, and $\epsilon \sim \mathcal{N}(0,I)$ is Gaussian noise. The diffusion UNet $\mu_\theta$ is adapted to handle the 8-channel latents, and the $\mathcal{C}_{\theta}$ is constructed to inject the RGB structural guidance, while keeping the VAE frozen. Let $x=[S_{1,\lambda},S_{2,\lambda}]$ denote the polarization targets and $c$ the RGB conditioning image proportional to $S_{0,\lambda}$. We adopt an $z_0$-prediction parameterization to train $\mu_\theta$ and $\mathcal{C}_{\theta}$ with the clean latent prediction loss
\begin{equation}
    \mathcal{L}_{\text{diff}}=
    \mathbb{E}_{x, c, \epsilon, t}
    \Big[\big\|\mu_\theta(z_t,c,t)-z_0\big\|_2^2\Big].
\label{eq:noise_stg1}
\end{equation}
To enforce physical consistency, we decode the predicted latent $\hat z_0=\mu_\theta(z_t,c,t)$ to obtain ($\hat{S}_{1,\lambda}$, $\hat{S}_{2,\lambda}$), and compute $\widehat{\mathrm{DoLP}}_\lambda$ and $\widehat{\mathrm{AoP}}_\lambda$ via Eq.~\eqref{eq:2}. Let $\Omega$ denote all pixels and $\Omega_\tau = \left\{\mathbf{y} \in \Omega \mid \mathrm{DoLP}_\lambda(\mathbf{y}) > \tau \right\}$ be the observability mask defined by a DoLP threshold $\tau$. We supervise linear Stokes components globally with
\begin{equation}
    \mathcal{L}_s= \textstyle
    \frac{1}{|\Omega|}
    \sum_{\mathbf{y} \in \Omega}
    \Big( \big| \hat{S}_{1,\lambda}(\mathbf{y}) - S_{1,\lambda}(\mathbf{y})\big|+
    \big| \hat{S}_{2,\lambda}(\mathbf{y}) - S_{2,\lambda}(\mathbf{y}) \big| \Big),
\label{eq:stokes_loss}
\end{equation}
and constrain AoP only on $\Omega_\tau$, where $\Omega_\tau$ is computed from GT DoLP. This follows our observation in Fig.~\ref{fig:3} that AoP becomes ill-conditioned under weak polarization, while high-DoLP regions exhibit consistent AoP cues across channels. We define an angular distance $d_{\text{ang}}(\cdot,\cdot)$ that is $\pi$-periodic, and write
\begin{equation}
    \mathcal{L}_{\text{AoP}}= \textstyle
    \frac{1}{|\Omega_\tau|}
    \sum_{\mathbf{y} \in \Omega_\tau}
    d_{\text{ang}} \Big( \widehat{\mathrm{AoP}}_\lambda(\mathbf{y}),
    \mathrm{AoP}_\lambda(\mathbf{y}) \Big)
\label{eq:aop_loss}
\end{equation}
as AoP observability-aware loss. Thus, the physics loss is $\mathcal{L}_{\text{phys}} = \mathcal{L}_s + \gamma_{\text{AoP}} \mathcal{L}_{\text{AoP}}$, and the overall training objective for the first stage is
\begin{equation} \mathcal{L}_{\text{stg}_1}=\mathcal{L}_{\text{diff}}+\gamma_{\text{p}}\mathcal{L}_{\text{phys}}.
\label{eq:stg1_loss}
\end{equation}

\noindent \textbf{Training stage II}: In Stage II, we convert the multi-step diffusion process learned in Stage I into a one-step generator $\mathcal{G}_\theta$. This stage accelerates inference and, more importantly, enables stable end-to-end adaptation of the VAE encoder via LoRA: backpropagating through a long stochastic denoising trajectory is computationally expensive and often yields high-variance gradients for the encoder, which can lead to unstable updates. By distilling the diffusion model into a one-step mapping, we shorten the backpropagation path to a single denoising step, yielding a well-conditioned end-to-end generator that supports reliable gradient propagation while allowing task-specific physical supervision during distillation. We adopt distribution match distillation (DMD)~\cite{yin2024one} for this conversion, as it provides a distribution-level distillation objective by matching real/fake score functions, leading to more stable one-step training under our physics supervision. We equip the $\mathcal{E}_{vae}$ with LoRA adapters and denote the resulting encoder as $\mathcal{E}_\phi$. For each training sample, we obtain the clean latent and apply forward noising:
\begin{equation}
    \tilde{z}_{t^*} = \alpha_{t^*} \mathcal{E}_\phi(x) + \sigma_{t^*} \epsilon,
    \label{eq:diffusion2}
\end{equation}

\begin{algorithm}[!h]
\caption{Training Procedure for Stokes-Informed Stable Diffusion}
\label{alg:1}
\KwIn{
Paired dataset $\{(x,c)\}$ with $x=[S_{1,\lambda},S_{2,\lambda}]$ and $c\propto S_{0,\lambda}$;
}
\KwOut{One-step generator $\mathcal{G}_\theta$, and LoRA parameters $\phi$}

\BlankLine
\textbf{Stage I: Train multi-step diffusion (freeze VAE)}\;
Freeze VAE parameters (encoder/decoder), train UNet $\mu_\theta$ and ControlNet $\mathcal{C}_{\theta}$\;
\While{Stage I training}{
    Sample batch $(x,c)^B$\;
    $z_0 \leftarrow \mathcal{E}_{vae}(x)$ ($\mathbb{R}^{B \times 8 \times \frac{H}{8} \times \frac{W}{8}} \leftarrow \mathbb{R}^{2B \times 3 \times H \times W}$)\;
    Sample $t$ and $\epsilon\sim\mathcal{N}(0,I)$; obtain $z_t$ using Eq.~\eqref{eq:diffusion}\;
    $\hat{z}_0 \leftarrow \mu_\theta(z_t,c,t)$\;
    $(\hat{S}_{1,\lambda},\hat{S}_{2,\lambda}) \leftarrow \mathcal{D}_{vae}(\hat{z}_0)$ ($\mathbb{R}^{2B \times 3 \times H \times W} \leftarrow \mathbb{R}^{B \times 8 \times \frac{H}{8} \times \frac{W}{8}}$)\;
    Compute $\mathcal{L}_{\text{diff}}$ using Eq.~\eqref{eq:noise_stg1}, $\mathcal{L}_s$ using Eq.~\eqref{eq:stokes_loss} and $\mathcal{L}_{\text{AoP}}$ using Eq.~\eqref{eq:aop_loss}\;
    Form $\mathcal{L}_{\text{phys}}=\mathcal{L}_s+\gamma_{\text{AoP}}\mathcal{L}_{\text{AoP}}$ and $\mathcal{L}_{\text{stg1}}$ using Eq.~\eqref{eq:stg1_loss}\;
    Update $\mu_\theta$ and $\mathcal{C}_{\theta}$ by minimizing $\mathcal{L}_{\text{stg1}}$\;
}

\BlankLine
\textbf{Stage II: One-step distillation + LoRA fine-tuning}\;
Initialize $\mathcal{G}_\theta$ from $\mu_\theta$; Freeze $\mu_{\theta}$, $\mathcal{C}_{\theta}$, and $\mathcal{D}_{vae}$; Insert LoRA into the $\mathcal{E}_{vae}$\;
\While{Stage II training}{
    Sample batch $(x,c)^B$\;
    $\tilde{z}_0 \leftarrow \mathcal{E}_\phi(x)$ ($\mathbb{R}^{B \times 8 \times \frac{H}{8} \times \frac{W}{8}} \leftarrow \mathbb{R}^{2B \times 3 \times H \times W}$)\;
    Sample $t^*$ and $\epsilon\sim\mathcal{N}(0,I)$; obtain $\tilde{z}_{t^*}$ using Eq.~\eqref{eq:diffusion2}\;
    $\hat {\tilde{z}}_0 \leftarrow \mathcal{G}_\theta(\tilde{z}_{t^*},c,t^*)$\;
    Compute $g_{\text{dmd}}$ using Eq.~\eqref{eq:dmd}\;
    $(\hat{S}_{1,\lambda},\hat{S}_{2,\lambda}) \leftarrow \mathcal{D}_{vae}(\hat {\tilde{z}}_0)$ ($\mathbb{R}^{2B \times 3 \times H \times W} \leftarrow \mathbb{R}^{B \times 8 \times \frac{H}{8} \times \frac{W}{8}}$)\;
    Compute $\mathcal{L}_{\text{phys}}$ using Eq.~\eqref{eq:stokes_loss}--Eq.~\eqref{eq:aop_loss};
    Compute $\mathcal{L}_{\text{kl}}$ using Eq.~\eqref{eq:KL_loss}\;
    Update $\theta$ (in $\mathcal{G}_\theta$) and $\phi$ (LoRA in $\mathcal{E}_\phi$) using Eq.~\eqref{eq:stg2_update}\;
}
\end{algorithm}

\noindent where $(\alpha_{t^*},\sigma_{t^*})$ are taken from the Stage-I noise schedule at a fixed timestep $t^*$. We distill the Stage-I multi-step diffusion $\mu_\text{real}=\mu_\theta$ into a one-step generator $\mathcal{G}_\theta$ using DMD~\cite{yin2024one}, which minimizes an approximate KL divergence $\text{KL}(p_\text{fake}\,\|\,p_\text{real})$ via score differences.
Let $\hat{\tilde z}_0=\mathcal{G}_\theta(\tilde z_t,t^*,c)$ be the one-step denoised latent and $\hat{\tilde z}_t=\alpha_t\hat{\tilde z}_0+\sigma_t\epsilon$ its re-noised version under the same schedule; the KL gradient is estimated by the real/fake scores computed from the frozen $\mu_\text{real}$ and an online-trained auxiliary generator $\mu_\text{fake}$:
\begin{equation}
\textstyle
g_{\text{dmd}} =
\mathbb{E}_{t,c,\epsilon}\Big[
\big(s_{\text{fake}}(\hat{\tilde z}_t,t,c)-s_{\text{real}}(\hat{\tilde z}_t,t,c)\big)\cdot
\frac{\partial \hat{\tilde z}_t}{\partial \theta}
\Big].
\label{eq:dmd}
\end{equation}
where \(s_{\text{real}}(\hat {\tilde{z}}_t, t,c)\) and \(s_{\text{fake}}(\hat {\tilde{z}}_t, t,c)\) are computed by the $\mu_{\text{real}}$ and $\mu_{\text{fake}}$ using:
\begin{equation}
    \textstyle
    s_{\text{real}}(\hat {\tilde{z}}_t,t,c) = -\frac{\hat {\tilde{z}}_t - \alpha_t \mu_\text{real}(\hat {\tilde{z}}_t,t,c)}{\sigma_t^2}, \quad
    s_{\text{fake}}(\hat {\tilde{z}}_t,t,c) = -\frac{\hat {\tilde{z}}_t - \alpha_t \mu_\text{fake}(\hat {\tilde{z}}_t,t,c)}{\sigma_t^2}.
    \label{eq:score_functions}
\end{equation}
We keep $\mu_{\text{real}}$ frozen and train $\mu_{\text{fake}}$ online Eq.~\eqref{eq:noise_stg1}, while updating $\theta$ using Eq.~\eqref{eq:dmd}. We reuse the physics loss $\mathcal{L}_{\text{phys}}$ defined in Stage I after decoding the predicted latent, and noting that $\mathcal{L}_s$ also serves as a regression loss for distillation. Since LoRA adapters are inserted into the encoder $\mathcal{E}_\phi$, we regularize its posterior to ensure that the learned latent $z$ distribution aligns with the standard Gaussian prior $p(z) = \mathcal{N}(0, I)$, as in standard VAEs. Specifically, the encoder defines $q_\phi(z | x) = \mathcal{N}(m_\phi(x), \mathrm{diag}(v_\phi^2(x)))$ with prior $p(z) = \mathcal{N}(0, I)$, where $m_\phi/v_\phi$ denote mean/standard deviation, and the KL term of LoRA is
\begin{equation}
    \mathcal{L}_{\text{kl}} = \mathbb{E}_{x}\left[\mathrm{KL}(q_\phi(z|x) \, || \, p(z))\right].
\label{eq:KL_loss}
\end{equation}

\noindent In Stage II, we update $\theta$ and $\phi$ with learning rate $\eta_\theta$ and $\eta_\phi$ as follows:
\begin{equation} 
\theta \leftarrow \theta - \eta_\theta \left( g_{\text{dmd}} + \nabla_\theta \mathcal{L}_{\text{phys}} \right),
\quad
\phi \leftarrow \phi - \eta_\phi (\nabla_\phi \mathcal{L}_{\text{phys}}+\nabla_\phi \mathcal{L}_{\text{kl}}),
\label{eq:stg2_update}
\end{equation}

\noindent During Stage II, we jointly optimize the one-step generator parameters $\theta$ and the LoRA parameters $\phi$ inserted in the encoder ($\mathcal{D}_{vae}$ fixed). In practice, $g_{\text{dmd}}$ distills the multi-step denoising behavior into $\mathcal{G}_\theta$, $\mathcal{L}_{\text{kl}}$ regularizes the LoRA-adapted encoder posterior, and $\mathcal{L}_{\text{phys}}$ provides physical supervision that is applied for both $\theta$ and $\phi$ to improve the stability of DoLP/AoP estimation. The overall training is summarized in Algorithm~\ref{alg:1}.

\section{Experiments}

\subsection{Experimental Setup}
\textbf{Data preparation}. We train and evaluate on multiple polarization datasets spanning two real capture modalities and one hybrid set:
($i$) Rotating linear polarizer group.
This group includes Monno~\cite{morimatsu2021monochrome,monno2025pol}, Qiu~\cite{qiu2021linear}, PIDSR~\cite{zhou2025pidsr}, and Wen~\cite{wen2021sparse}, which share the acquisition protocol of RLP captures (363 samples). We additionally capture 235 low-noise scenes using the same rotating-polarizer setup, yielding a total of 598 samples in the RLP group.
($ii$) Division-of-focal-plane polarimeter group. This group includes PolarFree~\cite{yao2025polarfree}, CPIF~\cite{luo2025cpifuse}, and Jeon~\cite{jeon2024spectral}, all collected by DoFP cameras, totaling 2188 samples.
($iii$) RSP group. RSP~\cite{kurita2023simultaneous} is treated as a separate group containing a mixture of simulated and real measurements, totaling 2000 samples. Across all datasets, we compute Stokes components following Eq.~\eqref{eq:1} (or the equivalent processing provided by the dataset when the native measurement format differs), and use $c\propto S_{0,\lambda}$ as the conditioning input and $x=[S_{1,\lambda},S_{2,\lambda}]$ as supervision. Furthermore, we split each group independently into training and testing sets. For each group, we randomly select 50 samples for testing; the rest are used for training.

\noindent \textbf{Metrics}. We evaluate polarization cues computed from the predicted Stokes components. Given predicted $(\hat{S}_{1,\lambda},\hat{S}_{2,\lambda})$ and the conditional input $S_{0,\lambda}$, we compute $\widehat{\mathrm{DoLP}}_\lambda$ and $\widehat{\mathrm{AoP}}_\lambda$ via Eq.~\eqref{eq:2}.
We report PSNR/SSIM on $\widehat{\mathrm{DoLP}}_\lambda$ and MeanAE on $\widehat{\mathrm{AoP}}_\lambda$. Metrics are computed per color channel and averaged over $\lambda\in\{R,G,B\}$. AoP is $\pi$-periodic and unreliable under weak polarization; hence, we compute AoP-MeanAE only on $\Omega_{\tau_{eval}}=\{\mathbf{y}\mid \mathrm{DoLP}_\lambda(\mathbf{y})>\tau_{eval}\}$ (Sec.~3) using a $\pi$-periodic angular distance. For fair comparison, $\Omega_{\tau_{eval}}$ is constructed from GT DoLP and is shared across all methods. Following the analysis in Fig.~\ref{fig:2}, we fix $\tau_{eval}=0.05$ for all experiments as a practical trade-off between AoP reliability and spatial coverage: it yields a reasonable cross-channel AoP consistency while retaining $\sim$70$\%$ of pixels in $\Omega_{\tau_{eval}}$.

\noindent \textbf{Baselines}. We retrain all baselines on the same union training set and evaluate them on identical group-wise test splits: PolarAnything~\cite{zhang2025polaranything}, Restormer, and MAE from RGB-to-Polarization~\cite{lin2025rgb2pol}. All methods use the same RGB intensity proxy $c\propto S_0$ as input. Since PolarAnything predicts grayscale DoLP/AoP cues whereas GenPolar predicts channel-wise RGB Stokes components, we preserve each method's original output representation during training and convert all outputs to a common DoLP/AoP space for evaluation.

\noindent \textbf{Implementation details.} We use SD1.5 as the base model and optimize all trainable modules with AdamW, using betas $(0.9,0.999)$ and weight decay $10^{-3}$. Stage I trains $(\mu_\theta,\mathcal{C}_\theta)$ for 100 epochs with a learning rate of $4\times10^{-5}$, while Stage II jointly trains the one-step generator $\mathcal{G}_\theta$ and encoder LoRA parameters $\phi$ for 30 epochs with a learning rate of $10^{-5}$, keeping $\mu_\theta$ and $\mathcal{D}_{vae}$ fixed. We set $\gamma_{\text{p}}=\gamma_{\text{AoP}}=0.5$, LoRA rank to 4, patch size to $512$, and batch size to 1. $S_1$ and $S_2$ are directly used in $[-1,1]$, while $S_0\in[0,2]$ is scaled to $[-1,1]$. ControlNet is initialized from the diffusion UNet. Stage-I inference uses 20 steps, and training is performed on 4 A40 GPUs.

\begin{figure}[tb]
  \centering
  \includegraphics[width=\textwidth]{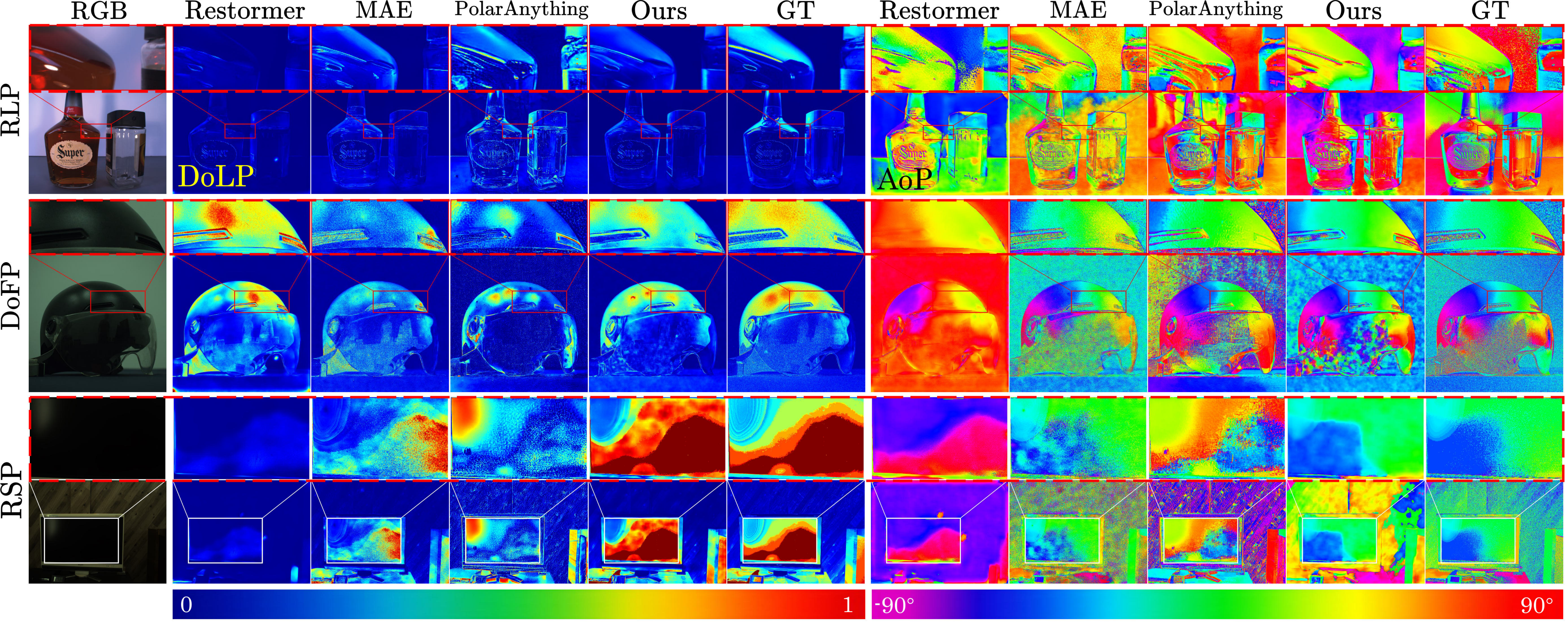}
  \caption{Visual comparisons among baselines and our GenPolar.}
  \label{fig:6}
\end{figure}

\subsection{Comparison with Prior Methods}

\noindent\textbf{Qualitative results.}
Fig.~\ref{fig:6} compares pseudo-colored DoLP and AoP predictions across methods. As PolarAnything only generates grayscale DoLP and AoP, pseudo-coloring is used for visualization consistency, and true-color results are provided in the supplementary material (SM). The first column shows the RGB input ($S_0$ proxy), followed by DoLP/AoP visualizations for all methods. GenPolar produces more coherent DoLP fields and AoP maps that better match the ground truth, with fewer artifacts and angular outliers. In contrast, PolarAnything tends to deviate from the GT in the zoomed DoFP and RSP regions, likely because direct DoLP/AoP prediction is more sensitive to AoP ambiguity in weakly polarized areas, while its channel-collapsed output provides less flexibility for wavelength-dependent polarization cues. Among deterministic baselines, Restormer improves upon MAE in some cases but is less robust to sensor artifacts, while MAE is stable but produces blurred DoLP boundaries and inconsistent AoP patterns. GenPolar’s performance benefits from two key factors: ($i$) an AoP observability-aware objective that stabilizes angles in weak-polarization areas, and ($ii$) LoRA-based VAE encoder adaptation, which reduces polarization-domain bias and stabilizes downstream DoLP/AoP.

\noindent \textbf{Quantitative results across modalities}.
Tab.~\ref{tab:main_compare_groups} reports DoLP-PSNR/SSIM and AoP-MeanAE on the three test groups. GenPolar achieves the best overall performance across capture modalities, with a clear advantage in AoP stability (lower MeanAE), especially on the DoFP and RSP groups where sensor artifacts and domain shifts make AoP reconstruction particularly challenging.

\begin{table}[t]
\centering
\setlength{\belowcaptionskip}{2pt}
\caption{Main comparison on polarization cues across test groups.}
\label{tab:main_compare_groups}
\resizebox{\linewidth}{!}{
\begin{tabular}{l|ccc|ccc|ccc}
\toprule
& \multicolumn{3}{c|}{RLP test} & \multicolumn{3}{c|}{DoFP test} & \multicolumn{3}{c}{RSP test}\\
Method
& PSNR $\uparrow$ & SSIM $\uparrow$ & MeanAE $\downarrow$
& PSNR $\uparrow$ & SSIM $\uparrow$ & MeanAE $\downarrow$
& PSNR $\uparrow$ & SSIM $\uparrow$ & MeanAE $\downarrow$ \\
\midrule
Restormer & 20.64 & 0.540 & 35.35 & 16.82 & 0.532 & 32.72 & 15.15 & 0.588 & 35.17 \\
MAE & 26.98 & 0.638 & 28.23 & 21.40 & 0.598 & 29.82 & 16.67 & 0.616 & 33.51 \\
PolarAnything & 19.77 & 0.563 & 33.56 & 17.85 & 0.572 & 29.33 & 15.47 & 0.604 & 30.45 \\
\midrule
GenPolar (Ours) & \textbf{29.54} & \textbf{0.741} & \textbf{20.15} & \textbf{24.01} & \textbf{0.666} & \textbf{18.77} & \textbf{21.29} & \textbf{0.676} & \textbf{27.60} \\
\bottomrule
\end{tabular}}
\end{table}

\subsection{Ablation Studies}

\begin{figure*}[t]
\centering
\begin{minipage}[t]{0.41\textwidth}
\captionsetup{type=table}
\captionof{table}{Quantitative ablations on the RLP testset.}
\label{tab:ablate_all}
\footnotesize
\centering
\setlength{\tabcolsep}{3pt}
\renewcommand{\arraystretch}{1.05}
\resizebox{\linewidth}{!}{
\begin{tabular}{lccc}
\toprule
Setting & PSNR $\uparrow$ & SSIM $\uparrow$ & MeanAE $\downarrow$ \\
\midrule
Only Stage I & 28.84 & 0.684 & 22.61 \\
Pre-LoRA & 27.92 & 0.674 & 23.49 \\
LoRA in Stage I & 19.78 & 0.460 & 50.03 \\
w/o LoRA & 28.49 & 0.681 & 23.12 \\
E2E + $\mathcal{L}_{\text{phys}}$ & 24.15 & 0.553 & 30.73 \\
w/o $\mathcal{L}_{\text{AoP}}$  & 27.21 & 0.668 & 23.96 \\
$\tau=0$  & 29.47 & 0.724 & 21.89 \\
Gray Stokes & 27.43 & 0.621 & 23.56 \\
$S_0$$\rightarrow$$[\mathrm{DoLP},\mathrm{AoP}]$ & 22.84 & 0.502 & 35.40 \\
\midrule
Ours & \textbf{29.54} & \textbf{0.741} & \textbf{20.15} \\
\bottomrule
\end{tabular}}
\par\smallskip
\captionsetup{type=figure,skip=2pt}
\includegraphics[width=\linewidth]{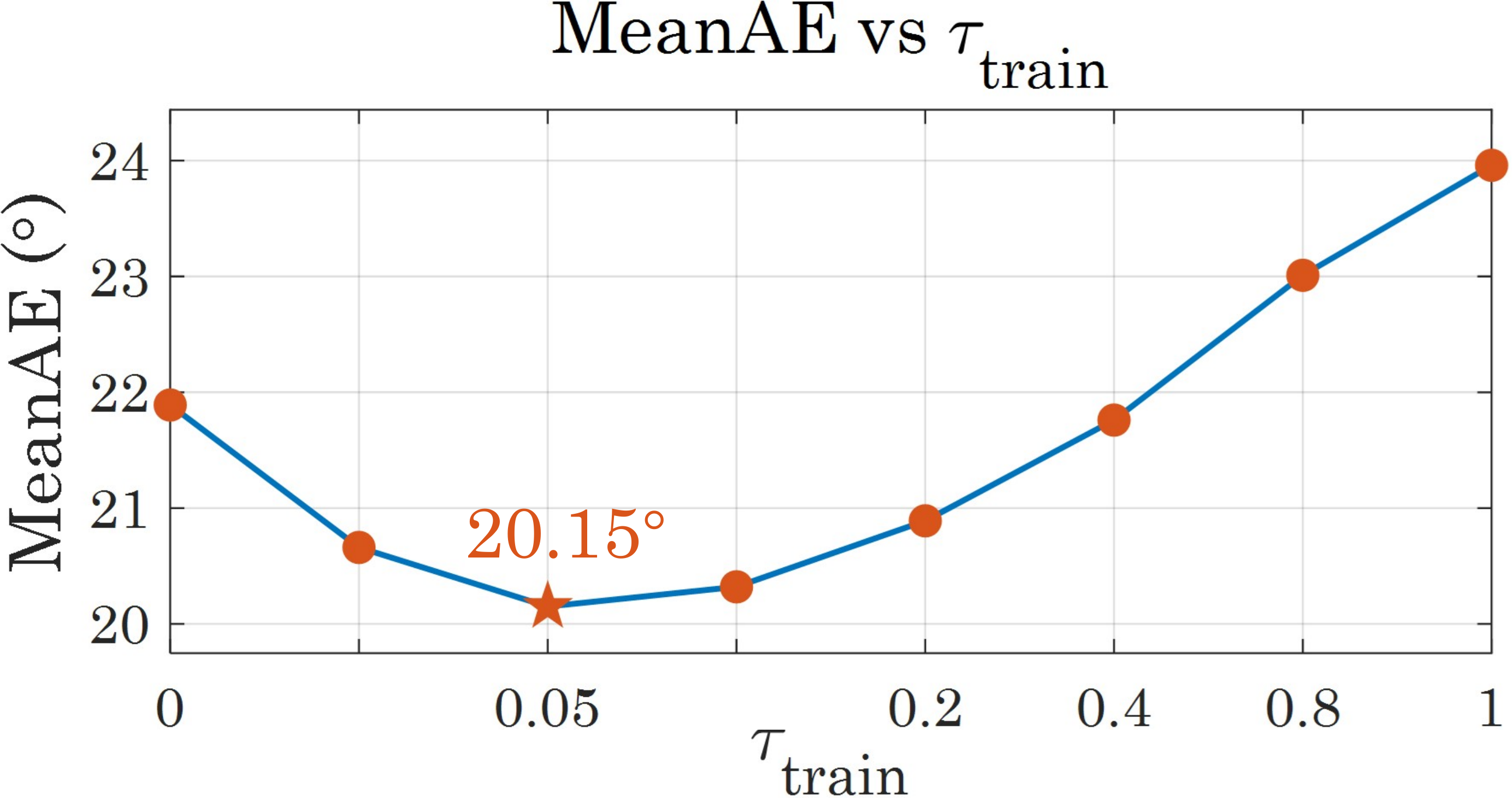}
\captionof{figure}{Ablation in $\tau_{train}$ of $\mathcal{L}_{\text{AoP}}$.}
\label{fig:7}
\end{minipage}\hfill
\begin{minipage}[t]{0.57\textwidth}
\centering
\raisebox{-\height}{\includegraphics[width=\linewidth]{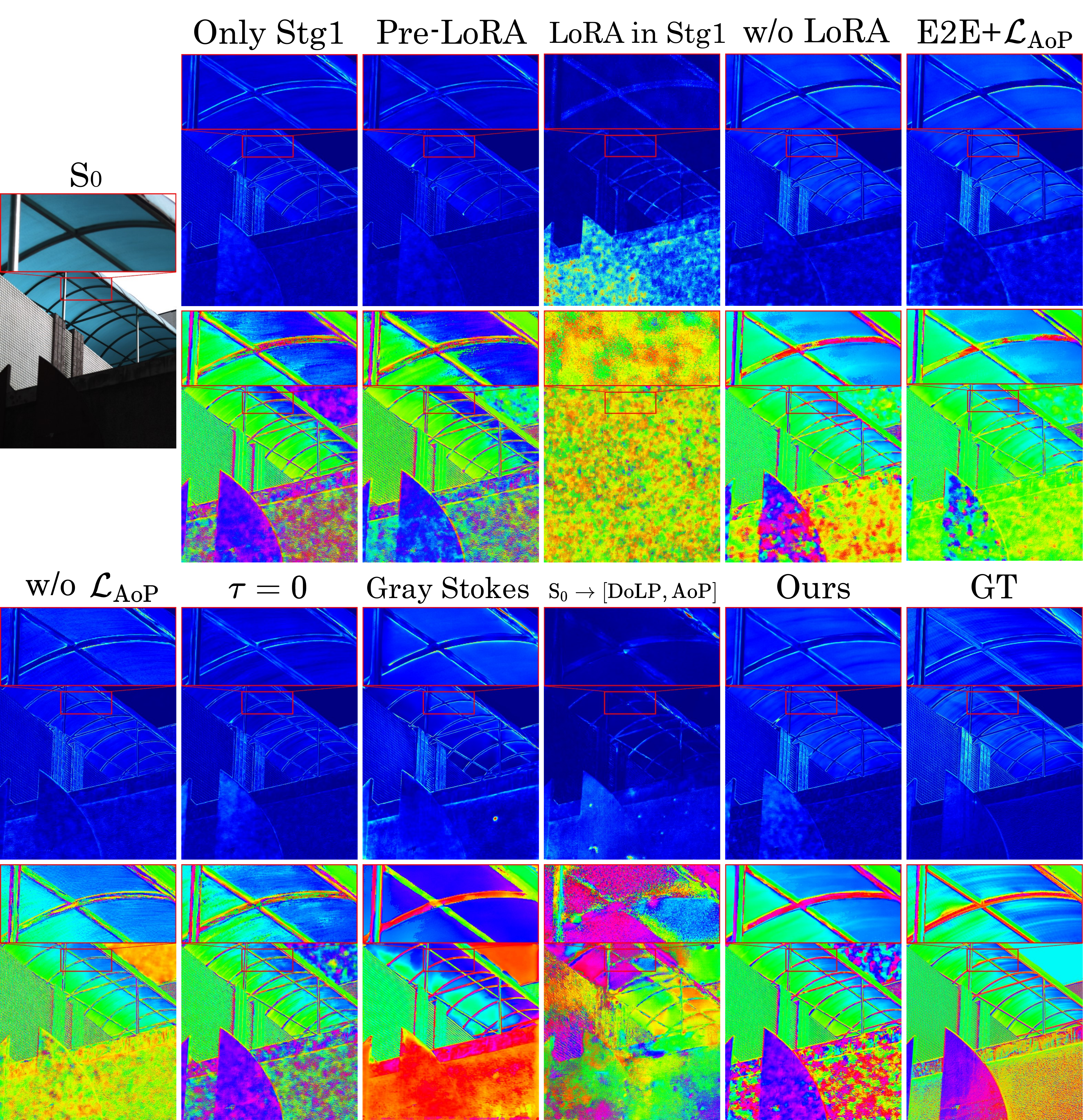}}
\captionsetup{type=figure}
\captionof{figure}{Visualization for ablation.}
\label{fig:8}
\end{minipage}
\end{figure*}

We ablate GenPolar on the RLP testset, which provides a controlled setting with consistent acquisition. As shown in Tab.~\ref{tab:ablate_all}, Fig.~\ref{fig:7} and~\ref{fig:8}, the quantitative metrics and visual examples are displayed. Detailed analysis is as follows.

\noindent \textbf{Stage II enables stable VAE adaptation}.
We report four ablations. ($i$) The Stage-I multi-step model with a frozen VAE already yields reasonable polarization cues. ($ii$) Pre-adapting $\mathcal{E}_{vae}$ with LoRA via reconstruction losses~\cite{rombach2022high} on our polarization data does not help either; using this adapted $\mathcal{E}_{vae}$ in Stage-I training further degrades the multi-step model, indicating that encoder adaptation is impractical under long diffusion trajectories. ($iii$) Directly inserting LoRA into $\mathcal{E}_{vae}$ during Stage-I multi-step training is unstable: DoLP remains plausible but AoP collapses. ($iv$) Distilling to a one-step generator resolves this optimization issue, but without LoRA the distilled model still inherits a latent-domain mismatch, leading to larger AoP error than the full model. Overall, one-step distillation is necessary to enable stable encoder adaptation, and LoRA is further needed to correct the latent mismatch for accurate AoP.

\noindent \textbf{Diffusion vs. E2E}.
For E2E + $\mathcal{L}_{\text{phys}}$, we replace diffusion sampling with a single feed-forward pass, using the diffusion UNet to directly map the RGB latent to the $(S_1,S_2)$ latent and training it end-to-end with $\mathcal{L}_{\mathrm{phys}}$. This baseline underperforms substantially, suggesting that the multi-step base diffusion provides useful distributional regularization beyond a single-pass regression network.

\noindent \textbf{Observability-aware AoP supervision.}
Removing $\mathcal{L}_{\mathrm{AoP}}$ ($\tau_{train}=1$) primarily harms AoP. The optimal AoP stability is achieved at $\tau_{train}=0.05$ according to Fig.~\ref{fig:7}, where the loss focuses on observable regions, preventing AoP from being enforced in weakly polarized areas.

\noindent \textbf{Channel-wise matters for color inputs.}
Gray-Stokes collapses $(S_{1,\lambda}, S_{2,\lambda})$ by replacing $(S_{1,R},S_{1,G},S_{1,B})$ with their channel mean (similarly for $S_2$), which degrades polarization cues and increases angular inconsistency. This supports our channel-wise Stokes formulation, which respects wavelength-dependent polarization strength and avoids the blurring effect introduced by channel averaging.

\noindent \textbf{Direct DoLP/AoP regression is unstable.}
Obvious degradation occurs for direct DoLP, $\sin(2\mathrm{AoP})$, and $\cos(2\mathrm{AoP})$ regression, since small sin/cos errors can translate into large angular deviations after inversion and this representation does not preserve Stokes-level coupling between magnitude and orientation.

Overall, the ablations show that robust linear polarization estimation benefits from jointly addressing representation, observability, and latent adaptation. Predicting per-channel $(S_{1,\lambda},S_{2,\lambda})$ avoids channel averaging that blurs wavelength-dependent polarization strength, while observability-aware AoP supervision  improves angular stability by avoiding ill-conditioned regions. Finally, the one-step generation training stage is essential for stable LoRA adaptation of the VAE encoder, yielding the strongest DoLP fidelity and AoP stability.

\begin{figure*}[t]
\centering
\begin{minipage}[t]{0.37\textwidth}
\captionsetup{type=table}
\captionof{table}{Material accuracy (dielectric vs.\ metallic).}
\label{tab:mat_det}
\small
\centering
\footnotesize
\setlength{\tabcolsep}{3pt}
\renewcommand{\arraystretch}{1.00}
\resizebox{0.7\linewidth}{!}{
\begin{tabular}{l c}
\toprule
Method & Acc. (\%) $\uparrow$ \\
\midrule
Captured & 93.75 \\
\midrule
MAE & 82.75 \\
Restormer & 75.12 \\
PolarA. & 45.36 \\
\midrule
Ours & \textbf{90.12} \\
\bottomrule
\end{tabular}}
\end{minipage}\hfill
\begin{minipage}[t]{0.61\textwidth}
\centering
\raisebox{-\height}{\includegraphics[width=\linewidth]{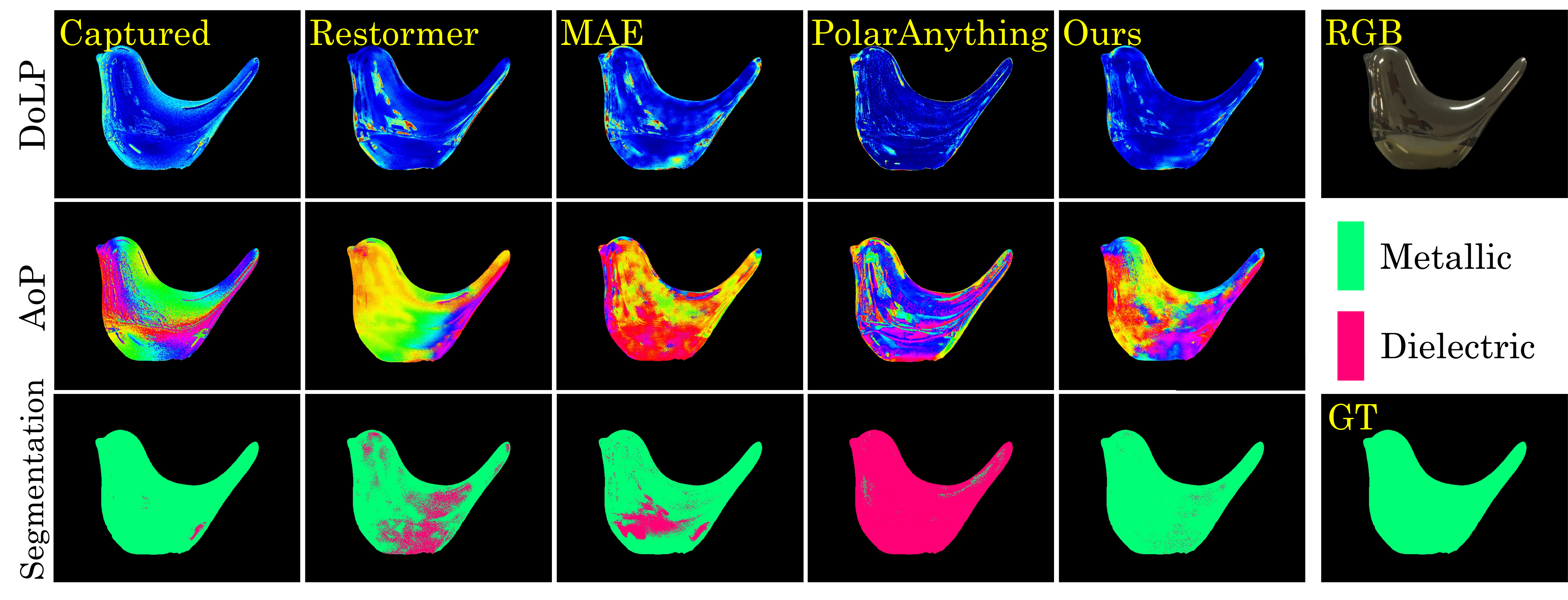}}\par
\captionsetup{type=figure}
\captionof{figure}{Qualitative material detection results (input DoLP/AoP and predicted masks).}
\label{fig:9}
\end{minipage}
\end{figure*}

\begin{figure*}[t]
\centering
\begin{minipage}[t]{0.37\textwidth}
\captionsetup{type=table}
\captionof{table}{De-reflection comparison in PSNR and LPIPS.}
\label{tab:deref}
\small
\footnotesize
\centering
\setlength{\tabcolsep}{3pt}
\renewcommand{\arraystretch}{1.05}
\resizebox{0.9\linewidth}{!}{
\begin{tabular}{l cc}
\toprule
Method & PSNR $\uparrow$ & LPIPS $\downarrow$ \\
\midrule
Captured & 22.44 & 0.133 \\
\midrule
MAE & 22.04 & 0.147 \\
Restormer & 21.87 & 0.169 \\
PolarA. & 20.08 & 0.190 \\
\midrule
Ours & \textbf{22.41} & \textbf{0.127} \\
\bottomrule
\end{tabular}}
\end{minipage}\hfill
\begin{minipage}[t]{0.61\textwidth}
\centering
\raisebox{-\height}{\includegraphics[width=\linewidth]{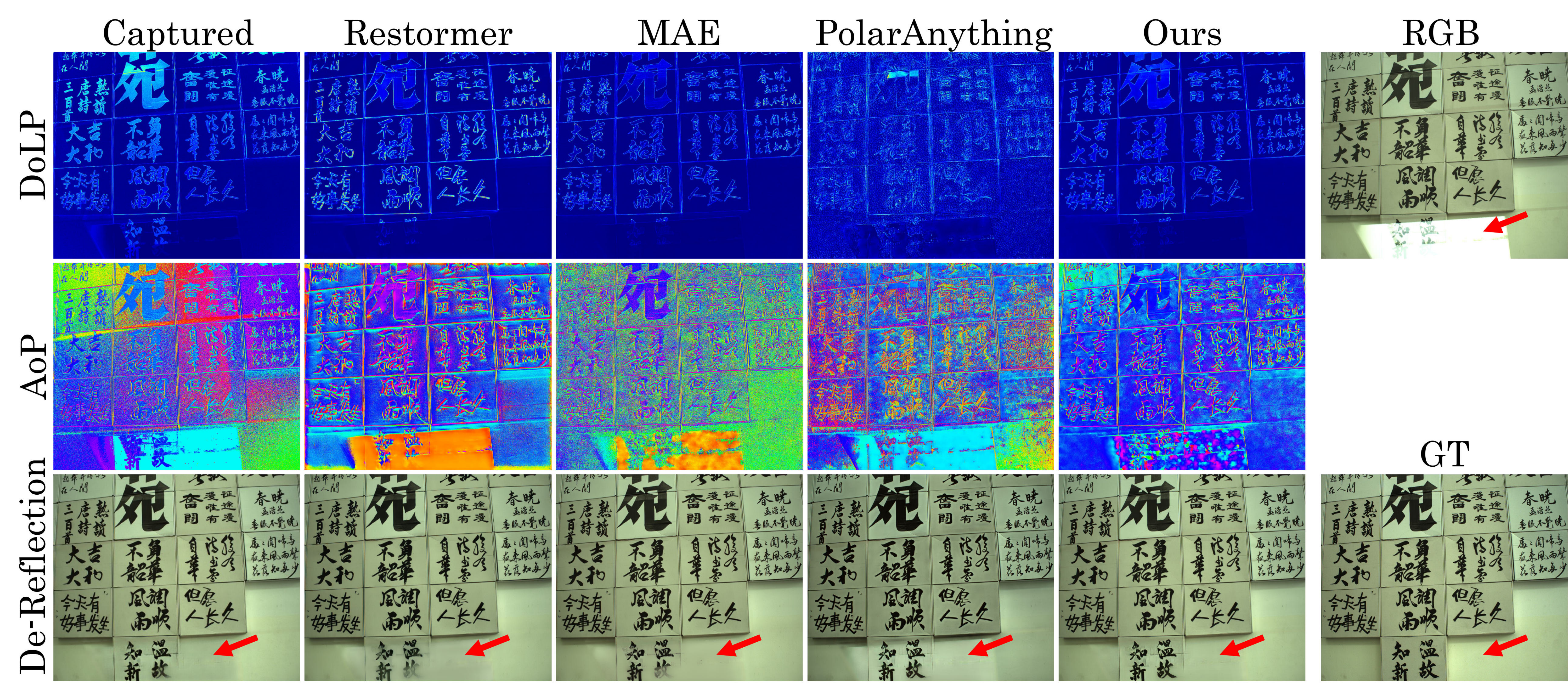}}\par
\captionsetup{type=figure}
\captionof{figure}{Visual input and de-reflected output.}
\label{fig:10}
\end{minipage}
\end{figure*}

\subsection{Downstream Applications}

We further evaluate whether the estimated polarization cues are usable for downstream pipelines, rather than only improving cue-level metrics. We consider two representative tasks with different output types: ($i$) material detection as a non-regression task (discrete categories), and ($ii$) polarization de-reflection as a regression task (continuous image restoration). In both tasks, we keep the downstream model fixed and only replace its polarization inputs (DoLP/AoP) with those computed from different methods, including polarization derived from measured captures and from four estimation baselines.

\noindent \textbf{Material detection}. We adopt SfPUEL~\cite{lyu2024sfpuel} to segment two material types (dielectric and metallic), and we evaluate detection accuracy on its real test set. For each method, we feed SfPUEL the corresponding DoLP/AoP cues, and obtain a material mask. Fig.~\ref{fig:9} visualizes the input DoLP/AoP and the resulting predictions. We observe that material discrimination is sensitive to both DoLP and AoP quality: specular reflections can corrupt polarization cues and lead to incomplete or fragmented masks. In particular, PolarAnything often confuses material categories and produces incorrect masks, while other baselines and even measured cues may suffer from reflection-induced artifacts. In contrast, GenPolar performs best among estimation methods and is close to the captured-input reference in both accuracy and visual quality, indicating that more coherent polarization cues improve downstream material reasoning.

\noindent \textbf{Polarization de-reflection}. We evaluate polarization-based de-reflection using PolarFree~\cite{yao2025polarfree}. Following the same protocol, we keep PolarFree fixed and only replace its polarization inputs with those from different methods. We report PSNR and LPIPS. Tab.~\ref{tab:deref} and Fig.~\ref{fig:10} shows that GenPolar consistently outperforms estimation baselines and remains close to the captured-input reference. Notably, PolarAnything is less catastrophic in this regression setting than in material detection, which is consistent with the visual results: it can partially suppress reflections but often leaves residual artifacts and loses fine details.

Across a non-regression material detection task and a regression de-reflection task, improved polarization cues translate to better downstream performance. These results serve as empirical evidence that observability-aware Stokes estimation can provide more usable DoLP/AoP cues for practical pipelines.

\subsection{Limitations}
GenPolar focuses on linear polarization, estimating $(S_1,S_2)$ and the derived DoLP/AoP, but does not recover the circular Stokes component $S_3$; full-Stokes estimation from RGB remains future work. Our effective unpolarized-illumination Mueller/Fresnel approximation may be less reliable under polarized illumination, skylight, transparent or multiple-scattering paths, and severe domain shift. These cases do not necessarily lead to failure, but they can weaken the RGB-to-polarization constraint; once DoLP is underestimated, AoP becomes less observable and its estimation accuracy inevitably degrades. Handling such coupled DoLP--AoP uncertainty remains an open problem. Representative failure cases are provided in the SM.

\section{Conclusion}

We presented GenPolar, a Stokes-informed stable diffusion framework for robust linear polarization estimation from a single RGB intensity proxy proportional to $S_0$. Our key insight is to predict channel-wise $(S_1,S_2)$ and treat AoP as observable only when DoLP is sufficiently high, incorporating this observability into both training and evaluation. To improve efficiency and reduce latent-domain mismatch, we adopt a two-stage pipeline that first trains a conditional diffusion model with physics-driven polarization supervision and then distills it into a one-step generator with lightweight encoder LoRA adaptation. Experiments across RLP, DoFP, and hybrid datasets show improved DoLP fidelity and AoP stability over retrained baselines, with consistent gains in downstream material detection and polarization-based de-reflection. Code is available \href{https://github.com/roydon-luo/GenPolar}{here}.

\section*{Acknowledgements}
This work was supported by the National Natural Science Foundation of China (grant number U2541205, 62271414),
National Key R\&D Program of China (2024YFF0505603), ``Pioneer” and “Leading Goose” R\&D Program of Zhejiang (grant number 2024SDXHDX0006, 2024C03182), the 2023 International Sci-tech Cooperation Projects under the purview of the “Innovation Yongjiang 2035” Key R\&D Program (grant number 2024Z126), the National Natural
 Science Foundation of China (grant number 62105372) and Hunan Provincial Research and Development Project (grant number 2025QK3019).

%
%
\bibliographystyle{splncs04}
\bibliography{main}

\clearpage
\appendix

\setcounter{section}{0}
\setcounter{equation}{0}
\setcounter{figure}{0}
\setcounter{table}{0}

\renewcommand{\thesection}{\Alph{section}}
\renewcommand{\theequation}{S\arabic{equation}}
\renewcommand{\thefigure}{S\arabic{figure}}
\renewcommand{\thetable}{S\arabic{table}}

\section{Details on DMD}

\begin{figure}[!h]
  \centering
  \includegraphics[width=\textwidth]{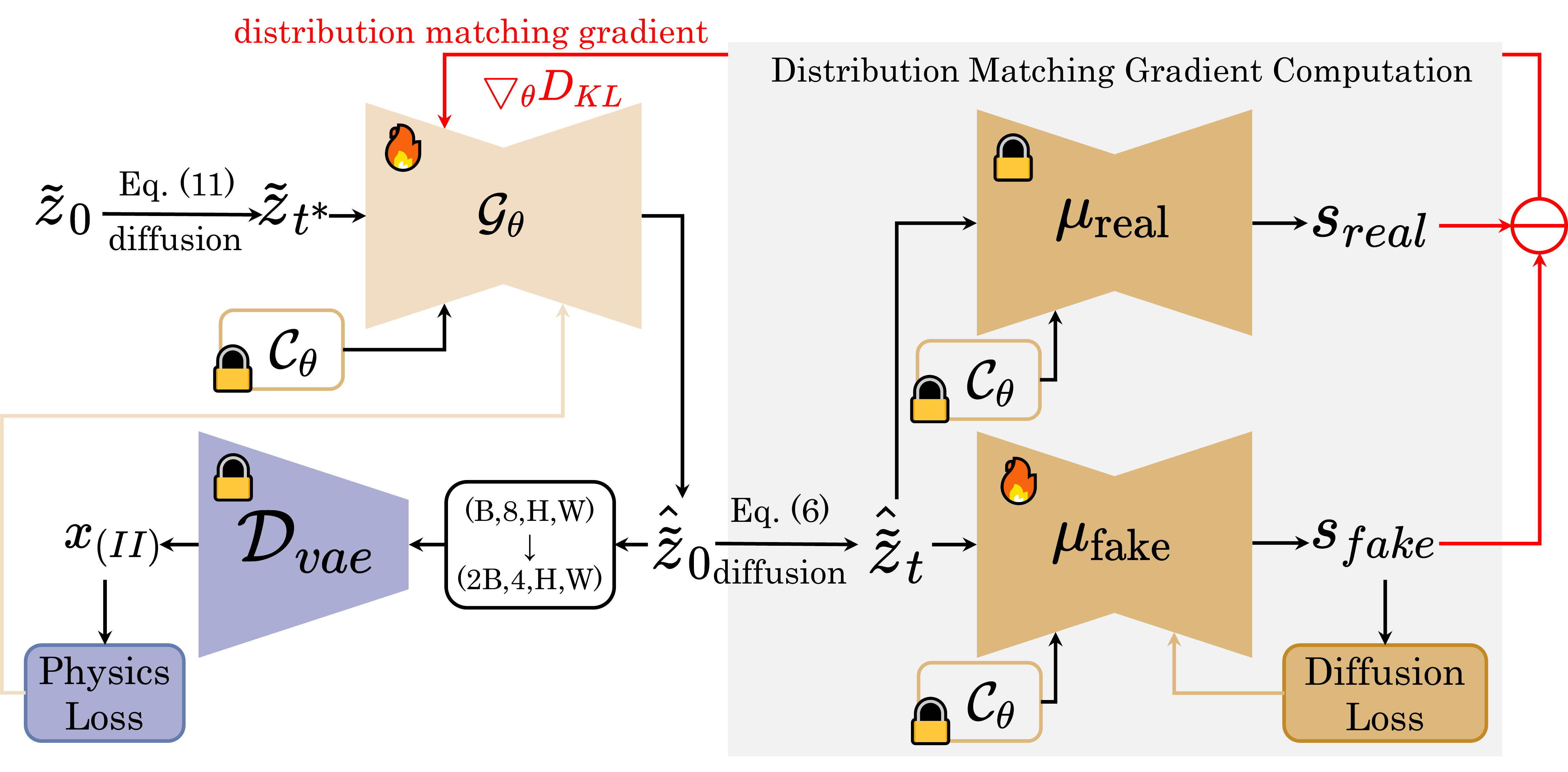}
  \caption{Overview of distribution matching distillation in our training stage II.}
  \label{fig:dmd}
\end{figure}

\noindent Fig.~\ref{fig:dmd} illustrates the Stage-II DMD mechanism. The only component not detailed in the main paper is the online update of the fake score model. Starting from the one-step output
\begin{equation}
    \hat{\tilde {z}}_0 = \mathcal{G}_\theta(\tilde z_{t^*}, c, t^*),
\end{equation}
we sample a random diffusion step $t$ and re-noise it as
\begin{equation}
    \hat{\tilde {z}}_t = \alpha_t \hat{\tilde {z}}_0 + \sigma_t \epsilon,\qquad \epsilon\sim\mathcal{N}(0,I).
\end{equation}
The frozen Stage-I diffusion model serves as the real score model ($\mu_{\mathrm{real}} = \mu_\theta$), while an auxiliary multi-step diffusion model $\mu_{\mathrm{fake}}$ is maintained online to estimate the score of the evolving generated distribution. Following the DMD formulation, the two scores are
\begin{equation}
s_{\mathrm{real}}(\hat{\tilde {z}}_t,t,c)
=
-\frac{\hat{\tilde {z}}_t-\alpha_t\mu_{\mathrm{real}}(\hat{\tilde {z}}_t,t,c)}{\sigma_t^2},
\qquad
s_{\mathrm{fake}}(\hat{\tilde {z}}_t,t,c)
=
-\frac{\hat{\tilde {z}}_t-\alpha_t\mu_{\mathrm{fake}}(\hat{\tilde {z}}_t,t,c)}{\sigma_t^2}.
\end{equation}
To keep $\mu_{\mathrm{fake}}$ synchronized with the current generator distribution, we update it online using the generated clean latent $\hat{\tilde z}_0$ as the denoising target:
\begin{equation}
\mathcal{L}_{\mathrm{fake}}
=
\mathbb{E}_{t,\epsilon}
\big[
\|
\mu_{\mathrm{fake}}(\hat{\tilde {z}}_t,t,c)-\hat{\tilde {z}}_0
\|_2^2
\big].
\label{eq:supp_fake_loss}
\end{equation}
Unlike the main one-step generator, $\mu_{\mathrm{fake}}$ is updated only in the latent domain and does not reuse the image-domain physics losses.
This simplification avoids an additional decoding path for the auxiliary model and thus reduces the memory and computational overhead of Stage II training.

The gradient flow in Stage II is separated as follows:
\begin{itemize}
    \item the DMD score-difference gradient updates only the one-step generator $\mathcal{G}_\theta$;
    \item the auxiliary latent diffusion loss $\mathcal{L}_{\mathrm{fake}}$ updates only $\mu_{\mathrm{fake}}$;
    \item the physics loss $\mathcal{L}_{\mathrm{phys}}$ updates both $\mathcal{G}_\theta$ and the LoRA parameters $\phi$ in the encoder;
    \item the posterior regularization term $\mathcal{L}_{\mathrm{kl}}$ updates only $\phi$;
    \item the real score model $\mu_{\mathrm{real}}$ and the VAE decoder remain fixed throughout Stage II.
\end{itemize}

\section{Additional analysis on LoRA}

To further analyze the role of LoRA, we compare three VAE adaptation strategies on the RLP test set:
(1) the original pretrained VAE without LoRA (\textsc{No-LoRA}),
(2) encoder LoRA pre-adapted only by plain autoencoding on the polarization training set (\textsc{Pre-LoRA}),
and (3) our Stage-II end-to-end LoRA adaptation (\textsc{Ours}). For each setting, we examine two aspects:
($i$) the direct VAE reconstruction behavior obtained by encoding and decoding the polarization target, and
($ii$) the final DoLP/AoP predictions after diffusion inference.
This comparison distinguishes improving autoencoding alone from improving the latent representation actually used by the diffusion model.

\begin{figure}[!t]
  \centering
  \includegraphics[width=\textwidth]{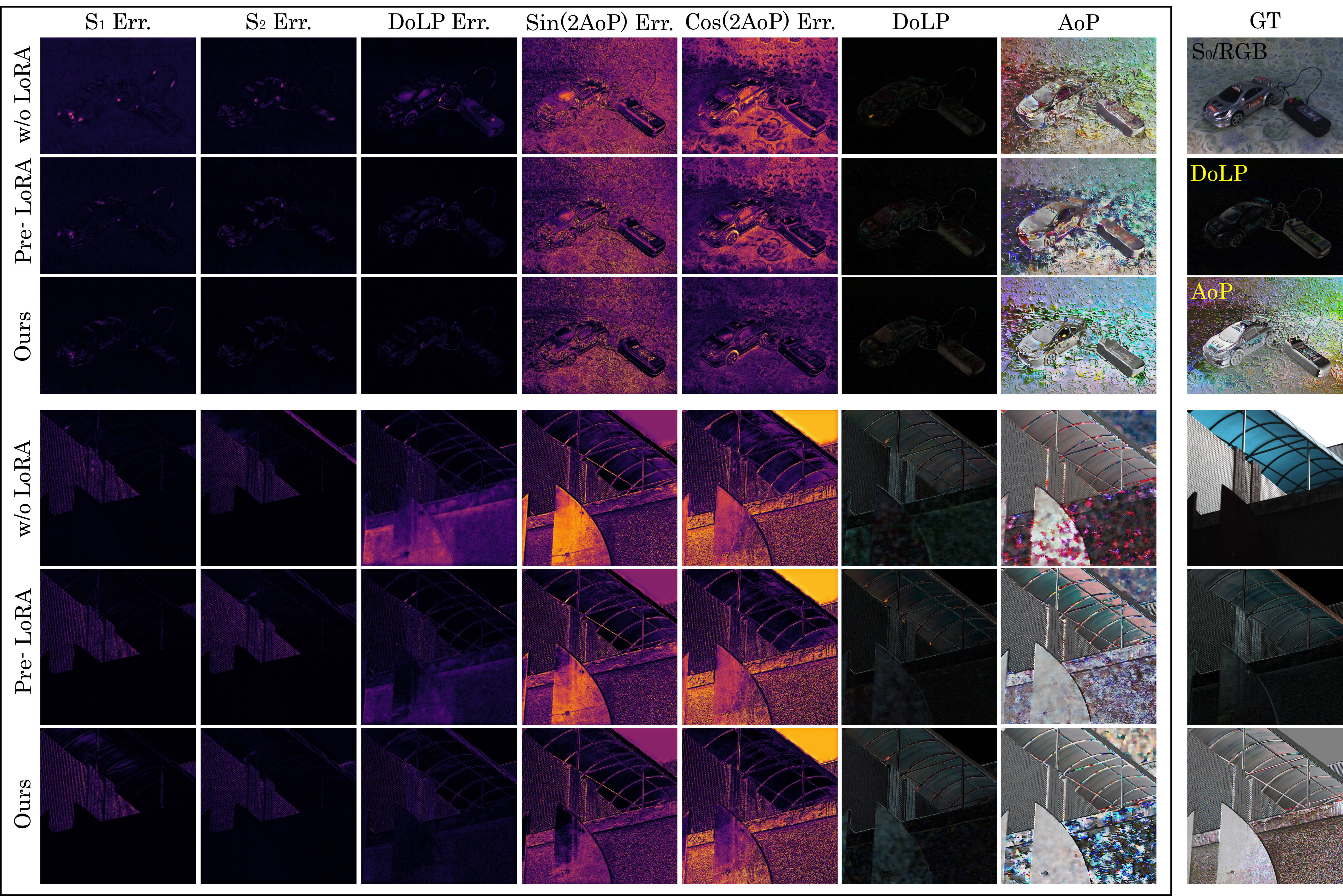}
  \caption{Error maps and DoLP/AoP outputs after diffusion across the three pipeline related to LoRA.}
  \label{fig:lora}
\end{figure}

As shown in Fig.~\ref{fig:lora}, the pretrained RGB VAE produces noticeable structured reconstruction errors on polarization targets, which propagate to the diffusion output and lead to relatively large DoLP/AoP errors.
The \textsc{Pre-LoRA} setting reduces the direct VAE reconstruction artifacts to some extent, indicating that encoder-only adaptation can partially fit the polarization dataset in an autoencoding sense.
However, this improvement does not translate to better DoLP/AoP after diffusion inference: the predicted cues remain similar to, or no better than, the \textsc{w/o-LoRA} case.
This shows that plain VAE pre-adaptation does not resolve the mismatch between the latent distribution expected by the diffusion UNet and the latent representation induced by the adapted encoder.

In contrast, our Stage-II end-to-end LoRA adaptation reduces the direct VAE reconstruction error and consistently improves the final diffusion outputs, producing more accurate DoLP/AoP maps with fewer artifacts.
This supports our main claim that LoRA is effective only when it is optimized jointly with the one-step generator under physics supervision, so that the adapted encoder remains aligned with the latent distribution used by the diffusion model.

\begin{table}[h]
\centering
\caption{
LoRA adaptation comparison on the RLP test set.
\textsc{Pre-LoRA} improves direct VAE reconstruction but does not reduce the final DoLP/AoP error after diffusion inference,
whereas our end-to-end LoRA improves both.
}
\begin{tabular}{l|ccccc|cc}
\hline
\multirow{2}{*}{Setting}
& \multicolumn{5}{c|}{VAE Recon. Mean Abs. Err. $\downarrow$}
& \multicolumn{2}{c}{Diffusion Output } \\
& $S_1$ & $S_2$ & DoLP & sin(2AoP) & cos(2AoP) & $\mathrm{PSNR_{DoLP}}$ $\uparrow$ & $\mathrm{MeanAE_{AoP}}$$\downarrow$ \\
\hline
\textsc{w/o LoRA}   & 0.020 & 0.019 & 0.081 & 0.476 & 0.503 & 28.49 & 23.12 \\
\textsc{Pre-LoRA}  & 0.010 & 0.012 & 0.035 & 0.307 & 0.375 & 27.92 & 23.49 \\
\textsc{Ours}      & \textbf{0.005} & \textbf{0.008} & \textbf{0.018} & \textbf{0.232} & \textbf{0.268} & \textbf{29.54} & \textbf{20.15} \\
\hline
\end{tabular}
\label{tab:lora}
\end{table}

A further observation is that the direct VAE reconstruction errors are much smaller for $S_1$, $S_2$, and DoLP than for angular quantities such as $\sin(2\mathrm{AoP})$ and $\cos(2\mathrm{AoP})$.
This is expected because AoP is obtained through a nonlinear angular transform of $(S_1,S_2)$, and small Stokes-domain perturbations can induce disproportionately large changes in angle-related representations, especially under weak polarization.
Therefore, the visible AoP degradation does not imply that the Stage-I multi-step diffusion is fundamentally unreliable; rather, it indicates that the remaining latent-domain reconstruction bias is sufficient to destabilize angular cues, which motivates the additional error reduction enabled by Stage-II end-to-end LoRA adaptation.

Tab.~\ref{tab:lora} reveals a clear mismatch between direct VAE reconstruction quality and final diffusion performance. Although \textsc{Pre-LoRA} substantially reduces the direct reconstruction error of \(S_1\), \(S_2\), DoLP, and the AoP-related proxies, it does not improve the final diffusion output, yielding even lower DoLP PSNR and higher AoP MeanAE than \textsc{w/o LoRA}. In contrast, our end-to-end LoRA improves both the direct VAE reconstruction and the final polarization estimation, confirming that standalone VAE pre-adaptation does not align the latent representation with the diffusion UNet, whereas joint optimization in Stage II does.

\section{Inference efficiency analysis}
We further analyze the inference efficiency of four methods: Restormer, MAE, PolarAnything, and our method. Table~\ref{tab:efficiency} reports both runtime and peak GPU memory usage under the same input resolution and hardware setting. The four methods differ not only in architecture, but also in prediction target and inference protocol.

\begin{table}[t]
\centering
\caption{
Inference efficiency comparison.
``Time'' denotes the end-to-end inference time per sample, and ``Peak Memory'' denotes the maximum GPU memory recorded during inference.
For MAE, $S_1$ and $S_2$ are obtained by two separate generative checkpoints; hence the reported runtime includes two sequential forward passes, while peak memory is measured per pass rather than summed across both passes.
}
\label{tab:efficiency}
\resizebox{0.98\linewidth}{!}{
\begin{tabular}{l|l|c|c}
\hline
Method & Output representation & Time (s) $\downarrow$ & Peak Memory (MB) $\downarrow$ \\
\hline
Restormer & color $(S_1,S_2)$ (3$\rightarrow$6 channels) & 0.1081 & 3733 \\
MAE & color $(S_1,S_2)$, separate generative prediction of $S_1$ and $S_2$ & 2.3378$\times$2 & 7400 \\
PolarAnything & grayscale DoLP, $\cos(2\mathrm{AoP})$, $\sin(2\mathrm{AoP})$ & 2.2623 & 7330 \\
Ours & color $(S_1,S_2)$ in one step & 0.1181 & 7434 \\
\hline
\end{tabular}
}
\vspace{-2mm}
\end{table}

\noindent \textbf{Restormer}.
Restormer is a direct discriminative regressor that takes the 3-channel $S_0$ proxy as input and jointly outputs the 6-channel color Stokes components $(S_1,S_2)$. Its single-pass design leads to both very fast inference and the lowest peak memory among the compared methods.

\noindent \textbf{MAE}.
For efficiency analysis, we characterize MAE by its actual prediction protocol. In our implementation, MAE serves as a conditioning prior for the latent generative backbone, playing a role functionally similar to the guidance branch in ControlNet~\cite{zhang2023adding}. Specifically, instead of using a shallow CNN-based condition encoder, it uses an MAE-based encoder to extract structural features from $S_0$ and inject them into the generative model. Importantly, the final $S_1$ and $S_2$ are generated by two separate checkpoints rather than predicted jointly in a single pass. This separate-generation protocol is not an implementation-specific workaround, but follows the original generative formulation adopted in RGB-to-Polarization~\cite{lin2025rgb2pol}, where the polarization components are modeled as separate generation targets. Therefore, two sequential forward passes are naturally required to obtain the complete color Stokes output, while peak memory is measured per pass rather than summed across both passes.

\noindent \textbf{PolarAnything}.
PolarAnything feeds $S_0$ into a ControlNet-style generative backbone and predicts single-channel DoLP, $\cos(2\mathrm{AoP})$, and $\sin(2\mathrm{AoP})$. As a result, it estimates polarization from color input, but does not estimate color polarization: the output representation itself is grayscale and discards wavelength-specific polarization information. Its inference time and memory are both dominated by the diffusion/ControlNet backbone.

\noindent \textbf{Our GenPolar}.
Our method directly predicts color Stokes components $(S_1,S_2)$ in a single step. Compared with MAE and PolarAnything, one-step distillation substantially reduces inference time while preserving a color-aware Stokes representation. Although our peak memory remains relatively high due to the Stable-Diffusion--based backbone and VAE decoder, the overall runtime is the lowest among the generative methods and remains close to the direct-regression Restormer baseline.

Overall, the comparison highlights two points.
First, one-step distillation primarily improves inference speed rather than eliminating model-residency memory.
Second, compared with existing generative baselines, our method achieves a more favorable trade-off: it is substantially faster than MAE and PolarAnything, while jointly predicting full color Stokes components instead of either generating $S_1$ and $S_2$ separately or collapsing polarization into grayscale cues.

\section{Special Fresnel cases and failure analysis}

In the main paper, we explain why DoLP is naturally wavelength-dependent through the factor $\rho(\lambda,\theta,n(\lambda))$. A further question is why AoP is often observed to be much less wavelength-sensitive in ordinary cases, and under what conditions this approximation breaks down. Here we clarify these special cases and analyze several representative failures of GenPolar.

\subsection{Why AoP is usually weakly wavelength-dependent}

Recall the formulation in the main paper:
\begin{equation}
\begin{bmatrix}
S_{1,\lambda}\\
S_{2,\lambda}
\end{bmatrix}
=
\rho(\lambda,\theta,n(\lambda))\,S_{0,\lambda}
\begin{bmatrix}
\cos 2\psi\\
\sin 2\psi
\end{bmatrix}.
\label{eq:supp_fresnel}
\end{equation}
Under ordinary single-bounce dielectric reflection, $\rho(\lambda,\theta,n(\lambda))$ carries the wavelength dependence through the Fresnel amplitudes and refractive index, while the angle $\psi$ is mainly determined by the plane of incidence, i.e., by local geometry and viewing direction.
Therefore,
\begin{equation}
\mathrm{DoLP}_{\lambda}=|\rho(\lambda,\theta,n(\lambda))|,
\qquad
\mathrm{AoP}_{\lambda}=\psi,
\end{equation}
which implies that DoLP is spectrally variant, whereas AoP is often approximately wavelength-independent in common reflective scenes. In other words, wavelength mainly changes the \emph{strength} of linear polarization, but not its dominant \emph{orientation}.

\begin{figure}[!t]
  \centering
  \includegraphics[width=\textwidth]{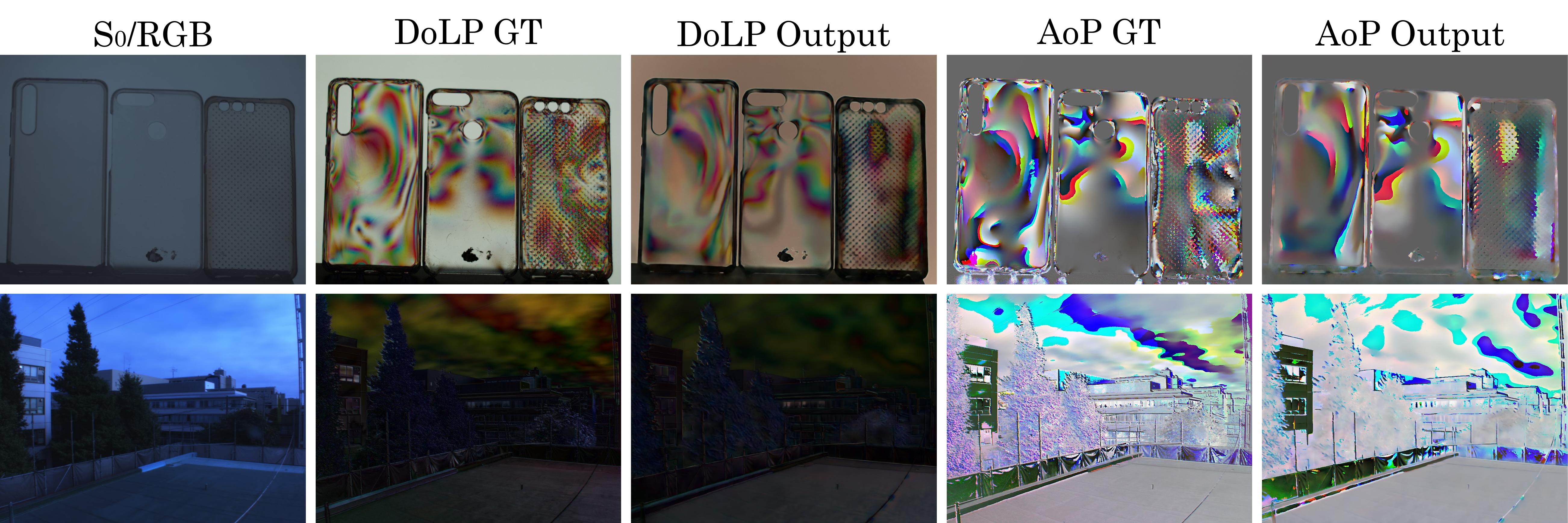}
  \caption{Special cases beyond the basic Fresnel interpretation. GenPolar remains effective on the transparent object, but the sky case is visibly more challenging.}
  \label{fig:spec}
\end{figure}

\subsection{Special cases: when AoP becomes wavelength-dependent}

The above approximation can break down when the effective polarization direction is no longer governed by a single reflection plane.
This is especially relevant for transparent media and sky regions.

\noindent\textbf{Transparent objects}.
For transparent or semi-transparent materials, the outgoing signal is often a mixture of reflection and transmission/refraction components.
In this case, the relevant direction inside the medium is governed by Snell's law~\cite{parazzoli2003experimental}:
\begin{equation}
n_1 \sin\theta_i = n_2(\lambda)\sin\theta_t(\lambda),
\end{equation}
which gives
\begin{equation}
\theta_t(\lambda)=\arcsin\!\left(\frac{n_1}{n_2(\lambda)}\sin\theta_i\right).
\end{equation}
Importantly, the external incident angle $\theta_i$ does not change with wavelength, but the refracted angle $\theta_t(\lambda)$ does because the refractive index is dispersive.
As a result, the internal propagation path, the relative contribution of different interfaces, and the effective polarization plane can all vary across RGB channels.
Consequently, the observed AoP may also exhibit noticeable wavelength dependence.
This explains why transparent objects, such as the transparent phone case in our example (Fig.~\ref{fig:spec}), can show channel-dependent AoP even though the simple single-reflection Fresnel picture would suggest $\mathrm{AoP}_{\lambda}\approx\psi$.

\noindent\textbf{Sky regions}.
For an ideal clear sky under the single-scattering Rayleigh model, the angle of polarization is perpendicular to the scattering plane and is theoretically independent of wavelength, while the degree of polarization varies spectrally~\cite{collett1992polarized}.
However, real captured sky images are rarely ideal Rayleigh-only observations, and are often affected by aerosols, clouds, and multiple scattering~\cite{pust2008digital}.
They often contain wavelength-dependent mixtures of molecular scattering, aerosols, multiple scattering, refraction, and even surface-reflected environmental light, all of which can perturb the effective polarization direction. Multiple scattering is known to modify the simple single-scattering polarization pattern, especially where single-scattering polarization is weak~\cite{collett1992polarized}.
Therefore, the sky example in our figure should not be interpreted as a contradiction to Rayleigh theory. Rather, it suggests that the captured sky region is not an ideal single-scattering case, and that additional effects in real scenes can cause the observed AoP to differ across channels.

Overall, these special cases suggest that the approximation $\mathrm{AoP}_{\lambda}\approx\psi$ is accurate for many ordinary reflective surfaces, but becomes less reliable when dispersive transmission, multi-interface mixing, or multiple scattering significantly contributes to image formation.
This also explains why our channel-wise formulation is still necessary even though AoP is often less spectral than DoLP.

\begin{figure}[!t]
  \centering
  \includegraphics[width=\textwidth]{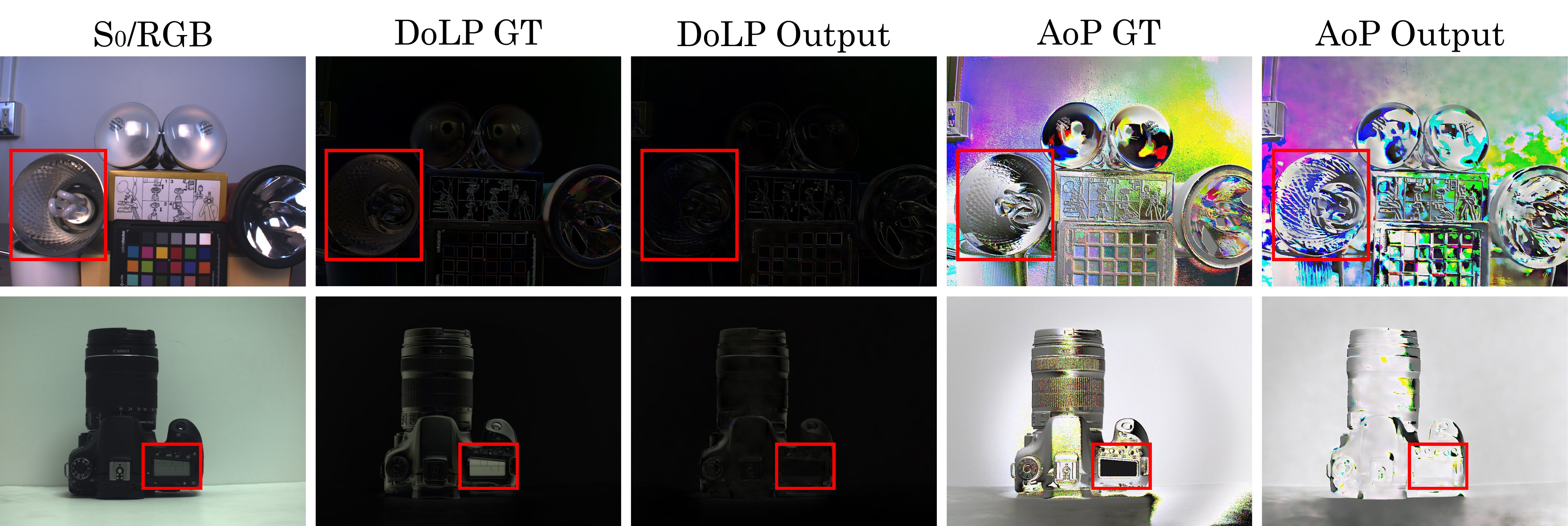}
  \caption{Representative failure cases of GenPolar. The red boxes mark regions where the GT indicates relatively reliable high-polarization areas. However, GenPolar underestimates the DoLP magnitude in these regions, and the weakened predicted linear polarization further causes large AoP deviations.}
  \label{fig:fail}
\end{figure}

\subsection{Failure cases of GenPolar}

We also present several failure cases of GenPolar in Fig.~\ref{fig:fail}.
In the red-boxed regions, the GT indicates that these pixels belong to relatively reliable high-polarization areas, where AoP should in principle be easier to estimate.
Nevertheless, our predictions still deviate noticeably from the GT.

A consistent observation is that the generated DoLP in these regions is lower than the GT.
This directly affects AoP estimation, because AoP is computed from the predicted Stokes components as
\begin{equation}
\widehat{\mathrm{AoP}}_{\lambda}
=
\frac{1}{2}\operatorname{atan2}(\hat S_{2,\lambda},\hat S_{1,\lambda}),
\end{equation}
and becomes highly sensitive to perturbations once the linear polarization magnitude is underestimated.
Although our training mask is constructed from GT DoLP, this only determines \emph{where supervision is applied}; it does not guarantee that the model prediction itself remains in a high-confidence regime.
Therefore, when the model underestimates the local polarization strength, the predicted DoLP drops below the GT level, and the corresponding AoP error increases accordingly. These examples reveal a practical gap between \emph{GT observability} and \emph{predicted observability}.
Our observability-aware loss prevents AoP from being enforced in clearly unreliable GT regions, but it cannot completely eliminate errors caused by insufficiently predicted Stokes magnitude in challenging cases.
This is consistent with the main-paper analysis: once the predicted polarization strength becomes weak, AoP is intrinsically more unstable.

\begin{figure*}[t]
    \centering
    \includegraphics[width=\textwidth]{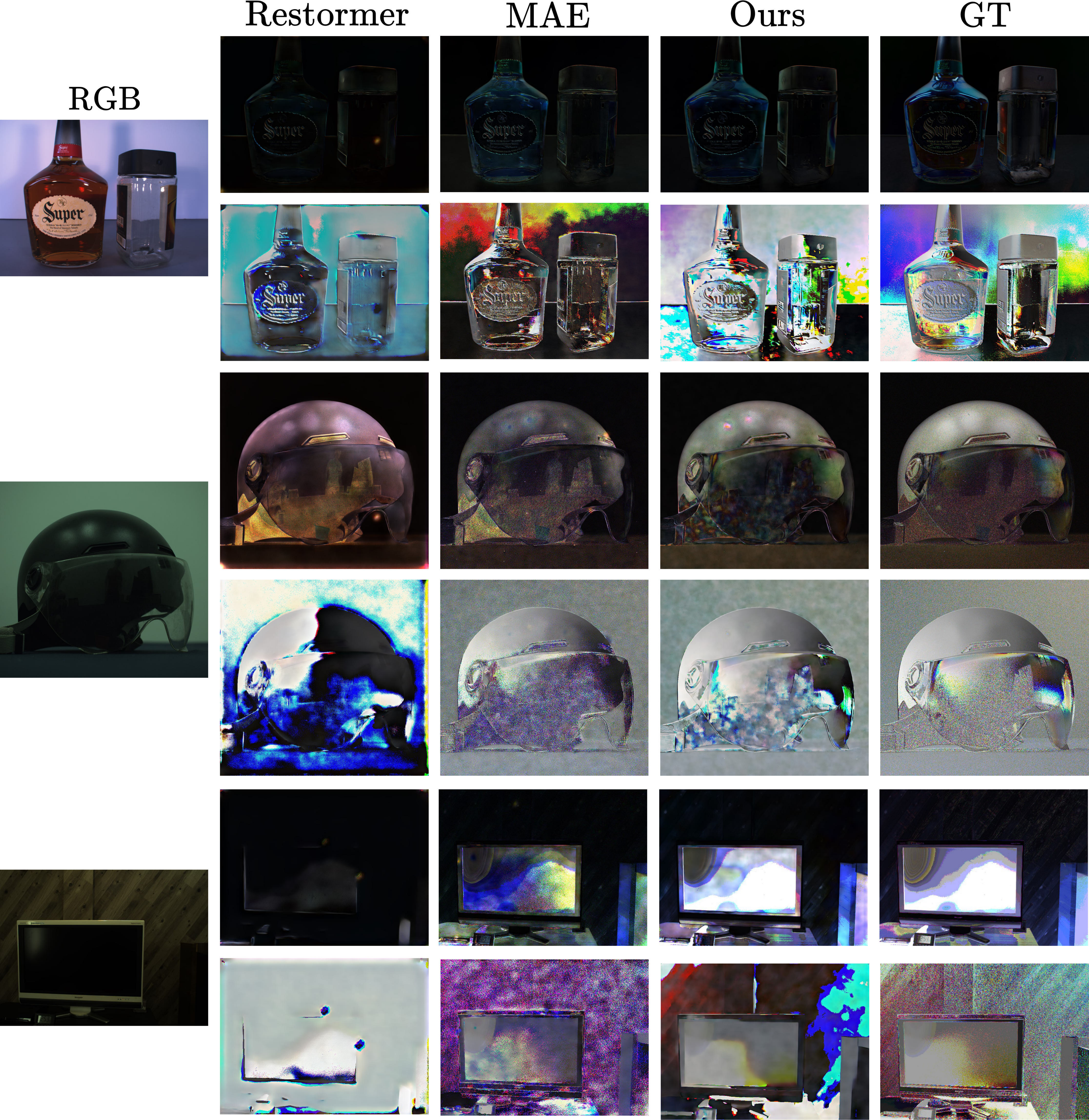}
    \caption{
    Additional true-color visualizations for the same three scenes shown in the main paper.}
    \label{fig:truecolor_main}
\end{figure*}

Qualitatively, we find that GenPolar still behaves reasonably well on many transparent objects, where local appearance cues remain informative for inferring the effective polarization transfer.
In contrast, sky reconstruction is more challenging, likely because its polarization formation depends on large-scale atmospheric scattering and spectral mixing that are only weakly constrained by local RGB appearance.

\begin{figure*}[t]
    \centering
    \includegraphics[width=\textwidth]{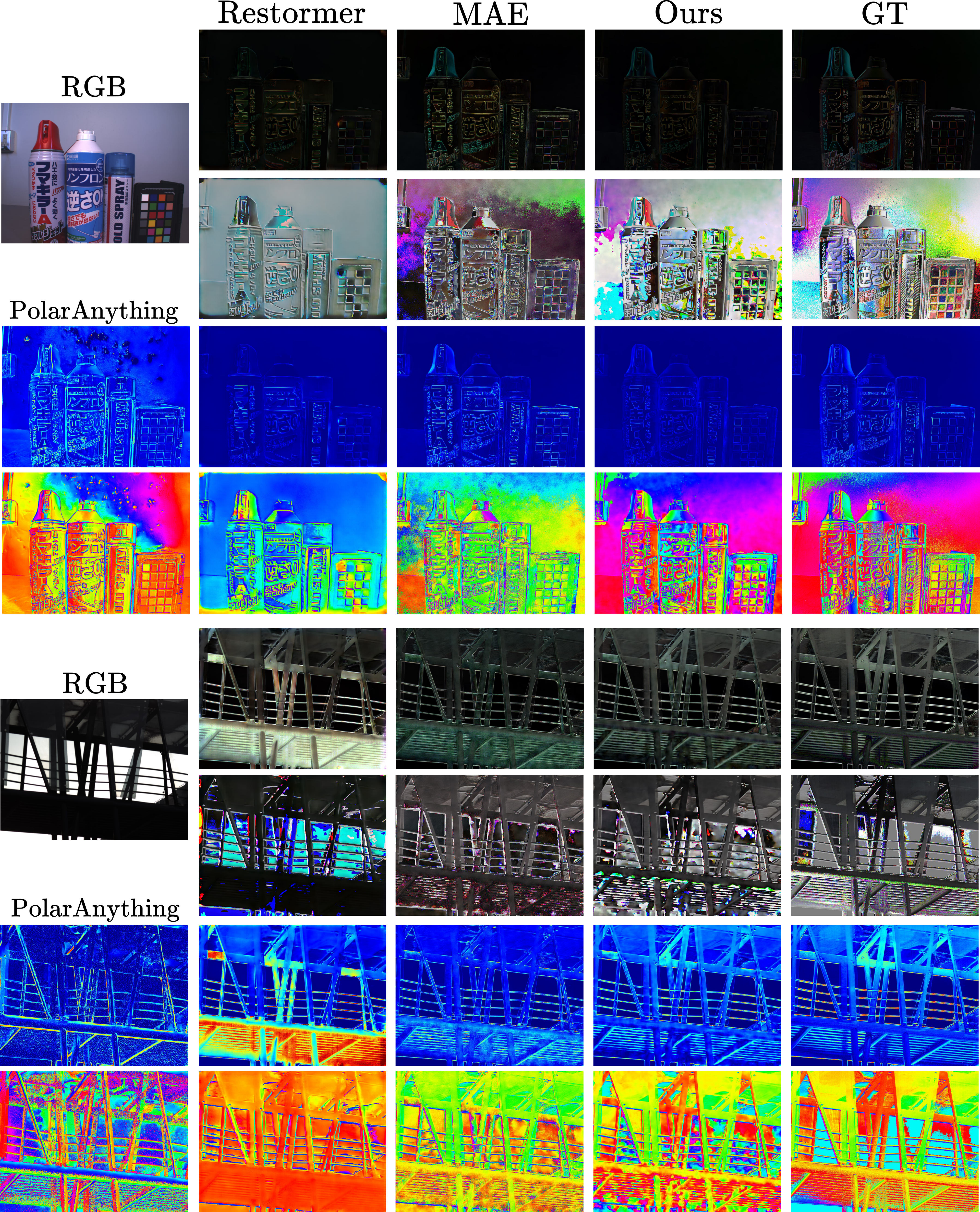}
    \caption{
    Additional qualitative results on the RLP test group. Two extra examples are shown, each with both pseudo-color and true-color visualizations.
    }
    \label{fig:more_rlp}
\end{figure*}

\begin{figure*}[t]
    \centering
    \includegraphics[width=\textwidth]{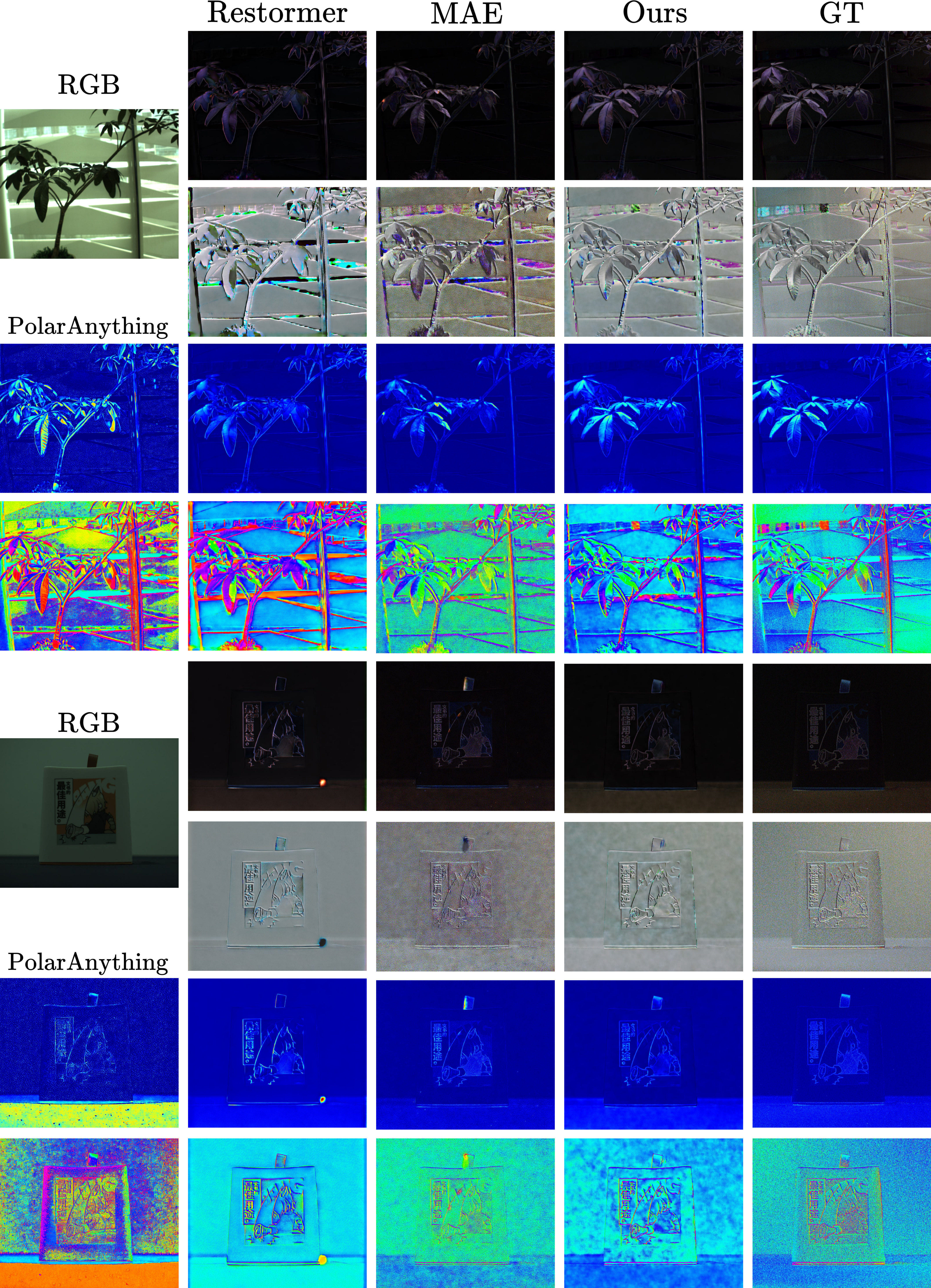}
    \caption{
    Additional qualitative results on the DoFP test group. Two extra examples are shown, each with both pseudo-color and true-color visualizations.
    }
    \label{fig:more_dofp}
\end{figure*}

\begin{figure*}[t]
    \centering
    \includegraphics[width=\textwidth]{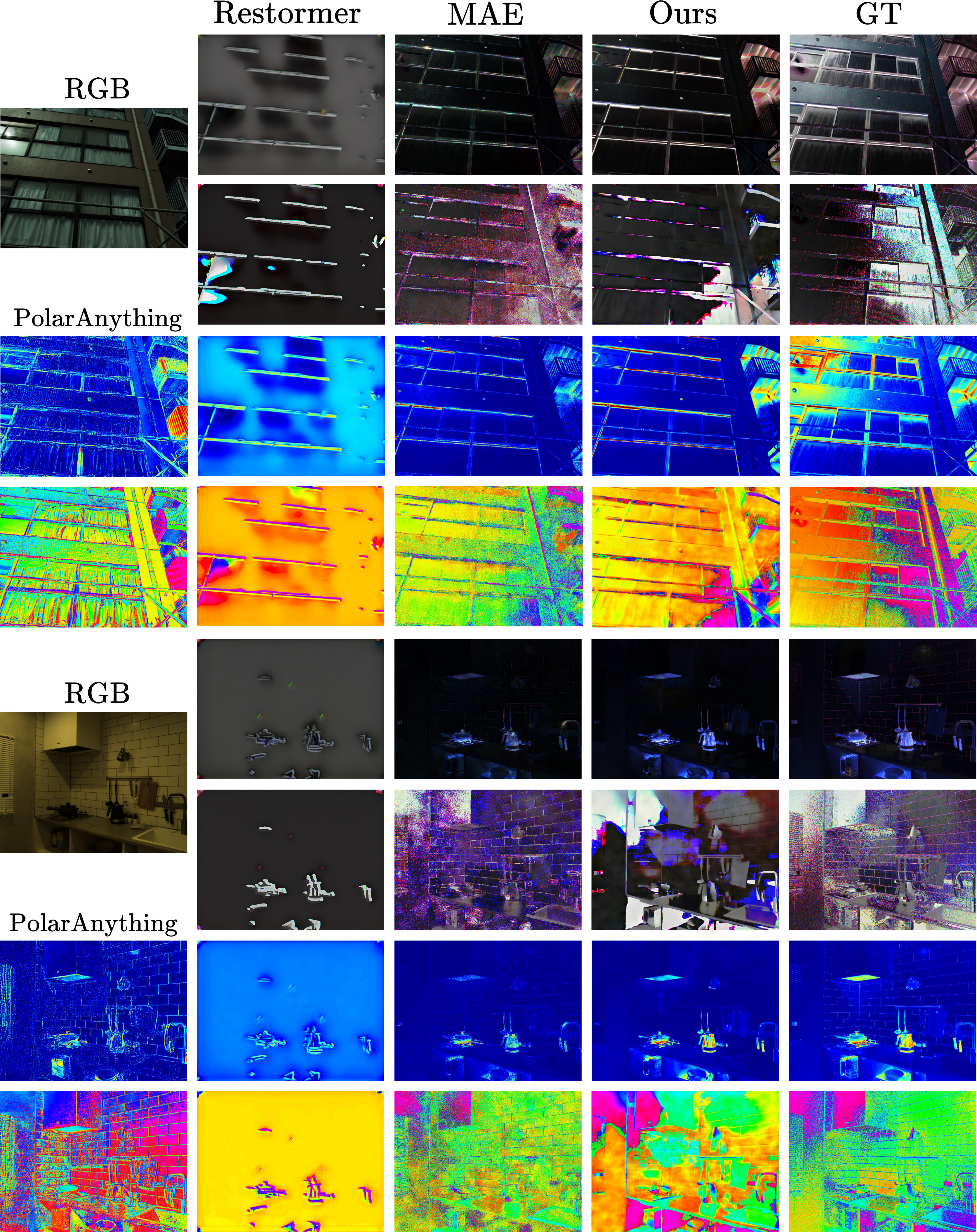}
    \caption{
    Additional qualitative results on the RSP test group. Two extra examples are shown, each with both pseudo-color and true-color visualizations.
    }
    \label{fig:more_rsp}
\end{figure*}

\section{Additional visual results}

\textbf{Additional true-color results}. The main paper mainly presents pseudo-color visualizations of DoLP and AoP for clearer structural comparison. Here we additionally provide true-color visualizations for the same three scenes in Fig.~\ref{fig:truecolor_main}, together with two extra examples for each of the three test groups in Figs.~\ref{fig:more_rlp}, \ref{fig:more_dofp}, and \ref{fig:more_rsp}. Compared with pseudo-color rendering, true-color visualization better reflects the channel-wise spectral characteristics of polarization.
Pseudo-color maps are useful for highlighting local structures and artifacts, but the visualization mapping may sometimes make a prediction look less consistent with the GT than it actually is. By contrast, true-color rendering more directly reflects whether the estimated polarization cues are spectrally consistent with the GT.

\noindent \textbf{Additional downstream results}. We further provide extra qualitative results for the two downstream tasks used in the main paper, namely material detection and polarization de-reflection.

\begin{figure*}[t]
    \centering
    \includegraphics[width=\textwidth]{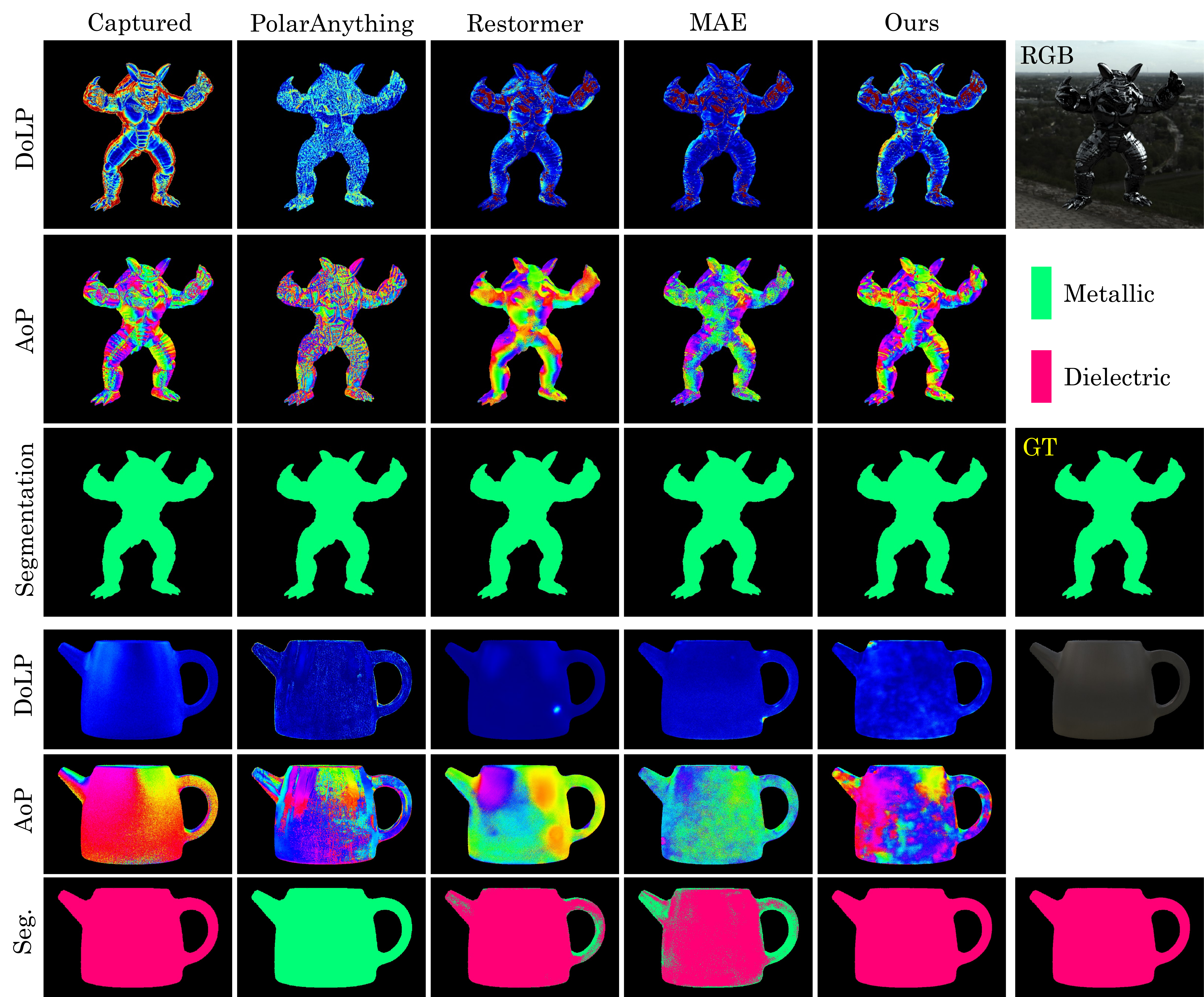}
    \caption{
    Additional qualitative results for material detection.
    Although the predicted AoP is not always visually perfect, the final material masks remain highly accurate, indicating that this task is more strongly driven by polarization magnitude cues.
    }
    \label{fig:more_material}
\end{figure*}

Fig.~\ref{fig:more_material} shows additional material-detection examples.
A notable observation is that the predicted AoP is not always visually perfect, yet the final material masks can still be highly accurate.
This suggests that, in our setting, material discrimination depends more strongly on whether the polarization \emph{strength} is correctly captured than on whether the polarization \emph{angle} is perfectly reconstructed.
Intuitively, the distinction between dielectric and metallic regions is more directly reflected by how strongly the surface polarizes light, whereas AoP is more sensitive to local geometry and thus can tolerate moderate errors without necessarily changing the final category prediction.
This explains why imperfect AoP estimation does not always harm the final detection result.

\begin{figure*}[t]
    \centering
    \includegraphics[width=\textwidth]{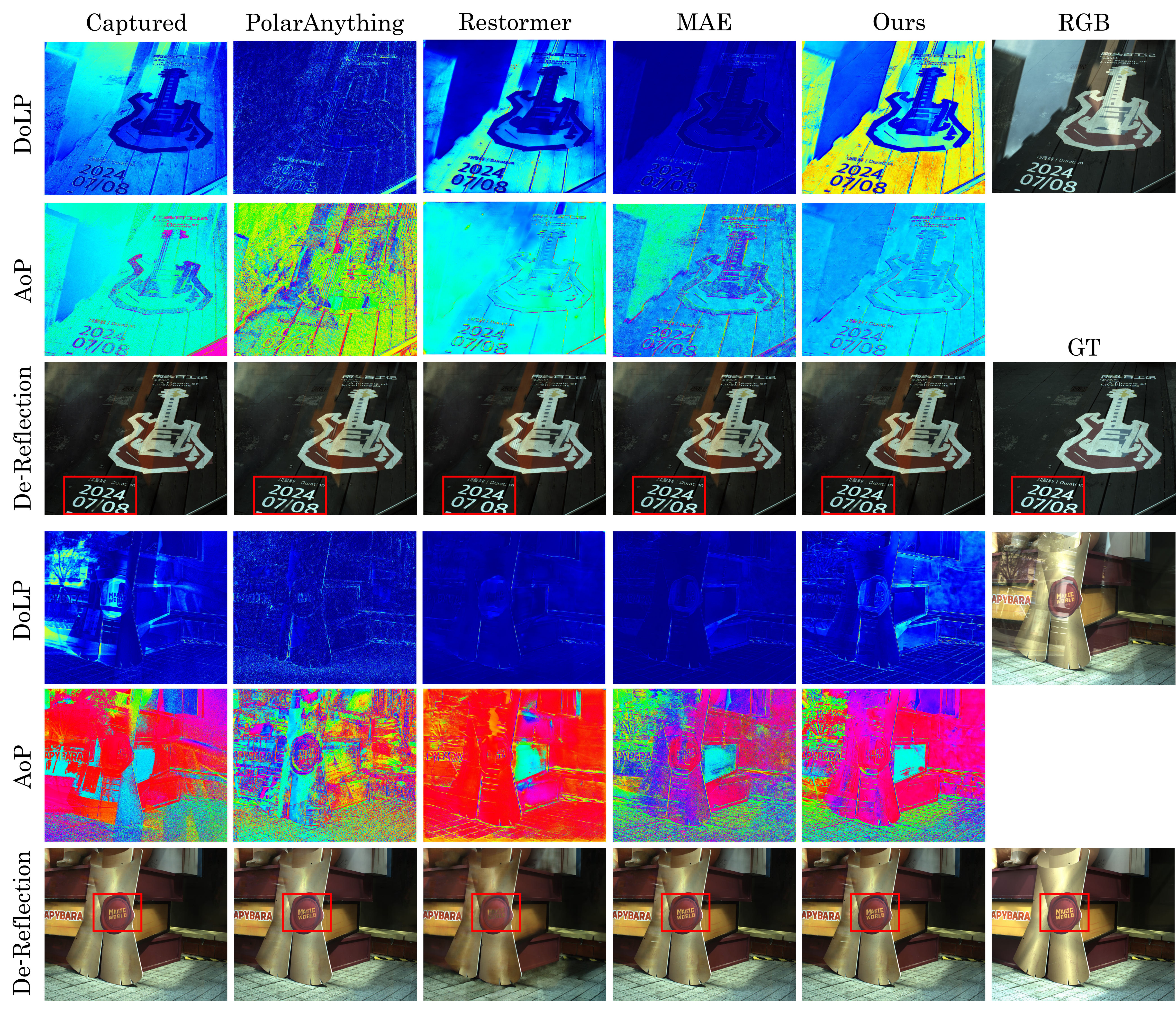}
    \caption{
    Additional qualitative results for polarization de-reflection.
    More accurate polarization estimation leads to cleaner reflection removal and better preservation of image details.
    }
    \label{fig:more_dereflection}
\end{figure*}

Fig.~\ref{fig:more_dereflection} shows additional de-reflection examples.
Compared with material detection, this task is more sensitive to the overall quality of the estimated polarization cues.
When DoLP and AoP are jointly reconstructed more coherently, the downstream method can better separate reflection from transmission, leading to cleaner reflection suppression and better preservation of fine image details.
The additional examples further support the same conclusion as in the main paper: improved polarization estimation translates to more reliable downstream restoration.

\end{document}